\documentclass[10pt,journal,compsoc]{IEEEtran}

\usepackage{times}
\usepackage{epsfig}
\usepackage{graphicx}
\usepackage{amsmath}
\usepackage{amssymb}
\usepackage[norule,symbol,perpage]{footmisc}
\usepackage{color}
\usepackage[colorlinks=true, citecolor={blue}, bookmarks=false, breaklinks]{hyperref}
\usepackage[numbers,sort&compress]{natbib}

\usepackage{multirow}
\usepackage{colortbl}
\usepackage{arydshln}

\usepackage{adjustbox}
\usepackage{url}

\usepackage{breakurl}
\usepackage{algorithm}
\usepackage[noend]{algpseudocode}
\newcommand{\R}{\mathbb{R}}

\begin{document}

\title{Master Face Attacks on\\Face Recognition Systems}

\author{Huy H. Nguyen,~\IEEEmembership{Member,~IEEE,}
        S\'ebastien Marcel,~\IEEEmembership{Senior Member,~IEEE,}\\
        Junichi Yamagishi,~\IEEEmembership{Senior Member,~IEEE,}
        and~Isao Echizen,~\IEEEmembership{Member,~IEEE}
\IEEEcompsocitemizethanks{\IEEEcompsocthanksitem H. Nguyen was with the Graduate University for Advanced Studies, SOKENDAI, Kanagawa, Japan.\protect\\
E-mail: nhhuy@nii.ac.jp
\IEEEcompsocthanksitem S. Marcel is with Idiap Research Institute, Switzerland.\\
J. Yamagishi is with the National Institute of Informatics and SOKENDAI, Japan.\\ 
I. Echizen is with the National Institute of Informatics, SOKENDAI, and University of Tokyo, Japan.
}
\thanks{Manuscript received December 1, 2020; revised 2021.}}

\markboth{Journal of \LaTeX\ Class Files,~Vol.~14, No.~8, August~2015}%
{Nguyen \MakeLowercase{\textit{et al.}}: A Study of the Master Face Attack on Face Recognition Systems}

\IEEEtitleabstractindextext{%
\begin{abstract}
	
Face authentication is now widely used, especially on mobile devices, rather than authentication using a personal identification number or an unlock pattern, due to its convenience. It has thus become a tempting target for attackers using a presentation attack. Traditional presentation attacks use facial images or videos of the victim. Previous work has proven the existence of master faces, \textit{i.e.}, faces that match multiple enrolled templates in face recognition systems, and their existence extends the ability of presentation attacks. In this paper, we perform an extensive study on latent variable evolution (LVE), a method commonly used to generate master faces. We run an LVE algorithm for various scenarios and with more than one database and/or face recognition system to study the properties of the master faces and to understand in which conditions strong master faces could be generated. Moreover, through analysis, we hypothesize that master faces come from some dense areas in the embedding spaces of the face recognition systems. Last but not least, simulated presentation attacks using generated master faces generally preserve the false-matching ability of their original digital forms, thus demonstrating that the existence of master faces poses an actual threat.
\end{abstract}
}

\maketitle

\IEEEdisplaynontitleabstractindextext
\IEEEpeerreviewmaketitle

\ifCLASSOPTIONcompsoc
\IEEEraisesectionheading{\section{Introduction}\label{sec:introduction}}
\else
\section{Introduction}
\label{sec:introduction}
\fi
Passwords should be strong, which can make them difficult to remember, and should be changed regularly to ensure security. Personal identification numbers (PINs) and unlock patterns are more convenient that passwords, but the user is still required to remember them, and people nearby may be able to steal a peek at them. An even more convenient method is biometric authentication, which uses a biometric trait unique to the user, eliminating the need to remember anything. This advantage has led to the widespread usage of biometric authentication on many portable devices including laptops and smartphones. The two most commonly used biometric traits for authentication are a fingerprint and the face. Since smartphones using this type of authentication may have a digital wallet (or e-wallet) for making e-payments, they are a prime target for attackers. An attacker may attempt to unlock such a device by performing a presentation attack~\cite{bhattacharjee2019recent}. For example, the attacker might attempt a presentation attack in which a printed facial image of the victim (known as a \textit{presentation attack instrument, or PAI}) is displayed in front of the smartphone's camera.

\begin{figure}[t]
	\centering
	\includegraphics[width=70mm]{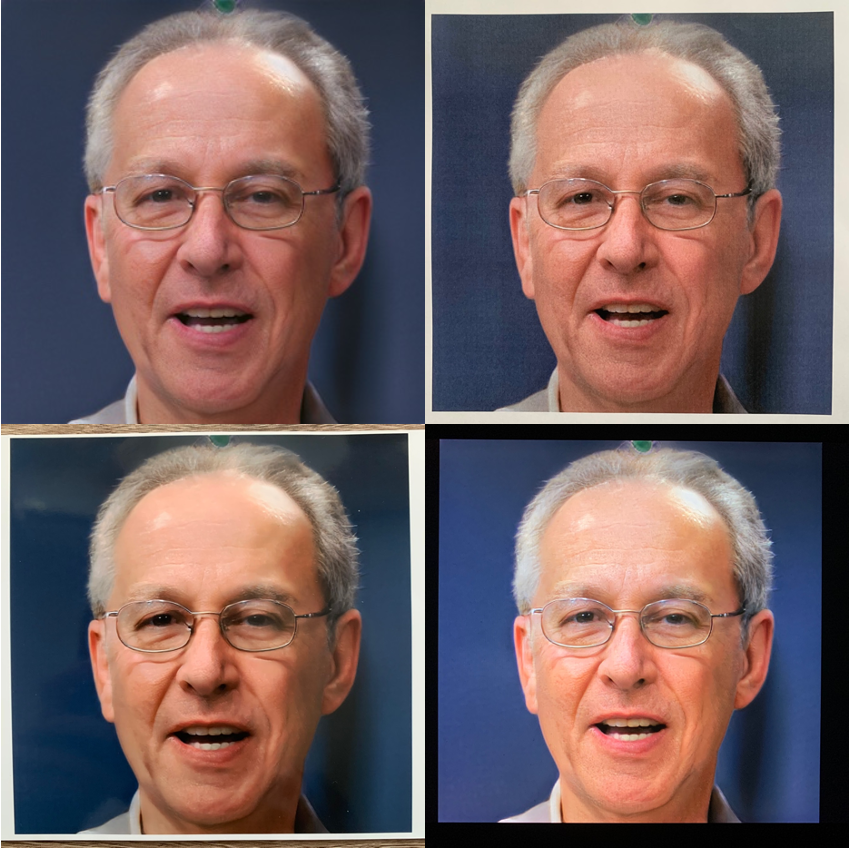}
	\caption{Original master face generated using two face recognition systems (top left) and its PAI forms printed on plain paper (top right), photo paper (bottom left), and displayed on a 13-inch Apple MacBook Pro screen (bottom right).}
	\label{fig:masterfaces}
\end{figure}

\begin{figure*}[t]
	\centering
    \includegraphics[width=165mm]{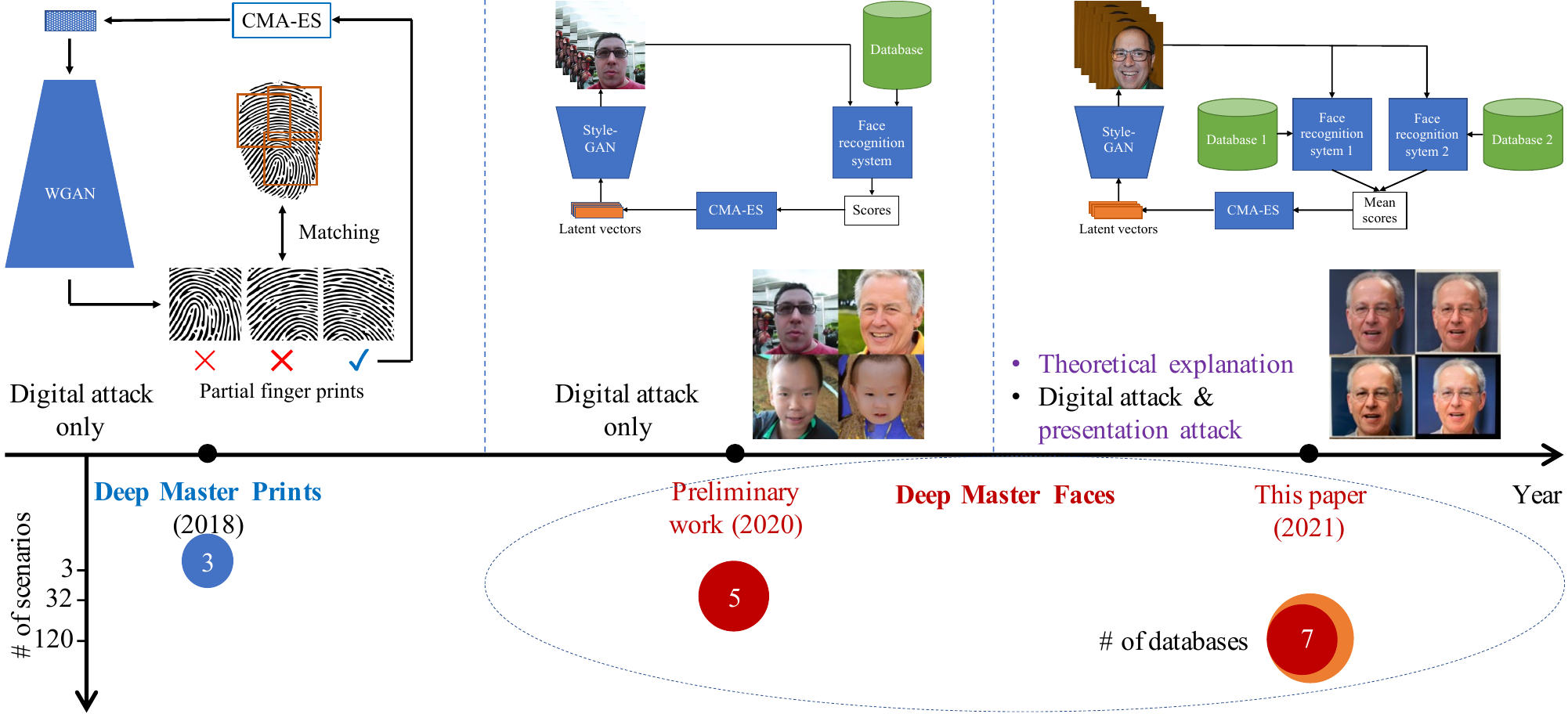}
	\caption{Stages in master biometrics research. First stage was partial master fingerprints as proposed by Bontrager \textit{et al.}~\cite{bontrager2018deep}. Next stage was our preliminary work on master faces~\cite{nguyen2020generating}. Current stage (this work) builds upon previous work and introduces extensions in algorithm, analysis, visualization, and test scenarios.}
	\label{fig:development}
\end{figure*}

The probability of a presentation attack succeeding becomes higher if the PAI matches multiple enrolled templates. In the facial domain, the creation of PAIs by blending together two or more faces is called face morphing~\cite{scherhag2019face}. The morphed face should match all source faces when used against a face recognition (FR) system and possibly even fool a human observer. This ability has made morphing a commonly used attack against automated border control systems in which the attacker ``borrows" the identity of the victim to enter or exit a location~\cite{scherhag2019face}. The face morphing approach is limited by the requirement that target faces be available. Another approach is to generate a ``master biometric" sample~\cite{bontrager2018deepmasterprints, nguyen2020generating}--a kind of ``wolf sample" that matches multiple enrolled templates in a biometric recognition system~\cite{une2007wolf}. This approach was first developed by Bontrager \textit{et al.}~\cite{bontrager2018deepmasterprints} for the fingerprint domain.
In our previous work~\cite{nguyen2020generating} and this extended work, we have adopted this approach and extended it to the facial domain. Unlike the face morphing approach, the attacker's advantage in this ``master face" approach is that it does not require any information about the victim. Moreover, using an ordinary PC and materials easily obtained from the Internet is enough to generate master faces concerning computational costs. However, before this work, the nature and characteristics of the master faces are not well (or sufficiently) understood.

The stages in master biometrics research are shown in Fig.~\ref{fig:development}. Our contributions can be summarized as follows:
\begin{itemize}
    \item This work, in combination with our previous work~\cite{nguyen2020generating}, is the first in which generated master faces can match multiple faces with different identities. This also implies FR systems are vulnerable to the master face attack.
    \item We extend our previous work by analyzing the effect of using multiple databases (DBs) and/or multiple FR systems for the latent variable evolution (LVE) algorithm to generate master faces. Some DB/FR system combinations boosted overall attack performance while others did not due to intra-component conflicts. Knowledge of the successful combinations is critical to understand in which conditions strong master faces could be generated and appropriately assess the potential risks.
    \item We expand the scope of our previous experiment by introducing more scenarios and using an additional facial database and an additional FR system trained with the angular margin loss~\cite{deng2019arcface} for the defender side. Furthermore, we introduce visualization in the face embedding (identity) space to obtain more insights into master faces. The gained insights are valuable to improve the robustness of the FR systems.
    \item To demonstrate the actual threat posed by the existence of master faces, we evaluated master face attacks by performing presentation attacks using printed images and the corresponding digital images displayed on a computer screen. Three PAIs are shown in Fig.~\ref{fig:masterfaces}.
\end{itemize}

The rest of the paper is organized as follows. First, we provide background information on facial image generation, FR systems, wolf attacks, master biometric attacks, and the LVE algorithm in section~\ref{sec:related_work}. Then, we discuss the existence of master faces and introduce an improved LVE algorithm using multiple databases and/or FR systems in section~\ref{sec:deep_master_faces}. Our experiments are covered in two sections: we first discuss generating master faces and their analysis in section~\ref{sec:generating}, and then discuss using master faces to perform presentation attacks in section~\ref{sec:presentation_attacks}. Next, in section~\ref{sec:defense}, we discuss ways to reduce the risk of master face attacks. Finally, we summarize the key points and make some closing remarks in section~\ref{sec:conclusion}.

\section{Related work}
\label{sec:related_work}
\subsection{Facial image generation}
Image generation is a major topic in deep learning research, and the face is a common target. There are two major approaches to image generation: using variational autoencoders (VAEs)~\cite{kingma2014auto} and using generative adversarial networks (GANs)~\cite{goodfellow2014generative}. In the beginning, they could only generate small images with low quality. VAEs tended to generate blurry images while GANs were difficult to train. Subsequent improvements in GANS (WGAN~\cite{arjovsky2017wasserstein} and WGAN Gradient Penalty (WGAN-GP)~\cite{gulrajani2017improved}) resolved the training problem, and GANs then began to be used to generate master prints~\cite{bontrager2018deepmasterprints}.

Recently improved versions of both VAEs~\cite{huang2018introvae, razavi2019generating} and GANs~\cite{brock2018large, karras2018progressive, karras2019style, karras2020analyzing} can generate high-resolution images. By gradually adding more layers during training to output larger images, Karras \textit{et al.} were able to generate $1024 \times 1024$ pixel images with their progressive GAN~\cite{karras2018progressive}. In subsequent work, they combined the ideas of progressive training and style transfer to create a better disentanglement network called StyleGAN~\cite{karras2019style}. Unlike traditional GANs, which directly use a latent vector for generating images, StyleGAN uses a mapping network to transfer this latent vector into intermediate style vectors used for synthesizing images. Controlling these intermediate style vectors changes the facial attributes. With the abilities of strong disentanglement and high-quality facial image generation, StyleGAN and its subsequent version~\cite{karras2020analyzing} are the best methods for generating master faces~\cite{nguyen2020generating}.

\subsection{Face recognition}
The release of large databases (\textit{e.g.}, the CASIA-WebFace database~\cite{yi2014learning} and the MS-Celeb database~\cite{guo2016ms}) and recent advances in convolutional neural networks (CNNs) have substantially improved the performance of FR systems and enabled them to work effectively in heterogeneous domains~\cite{de2018heterogeneous}. Most state-of-the-art FR systems~\cite{de2018heterogeneous, schroff2015facenet, deng2019arcface} make use of a network architecture that achieved high performance in the ImageNet Challenge~\cite{ILSVRC15}, such as the VGG network architecture~\cite{simonyan2014very} and the Inception network architecture~\cite{szegedy2017inception}. Parkhi \textit{et al.} trained the VGG-16 network on a custom-built large-scale database~\cite{parkhi2015deep} to create the VGG-Face network. Wu \textit{et al.} proposed a lightweight CNN that has ten times fewer parameters than the VGG-Face network~\cite{wu2018light}. The Inception architecture was used by de Freitas Pereira \textit{et al.} to build heterogeneous FR networks~\cite{de2018heterogeneous} and by Schroff \textit{et al.} to build the FaceNet network~\cite{schroff2015facenet}. Sandberg re-implemented FaceNet as an open-source system~\cite{sandberg2017facenet}. Taigman \textit{et al.} introduced DeepFace in which explicit 3D face modeling is used to improve the facial alignment phase and a CNN is used to extract face representation~\cite{taigman2014deepface}. Unlike previous methods, which use discriminative classifiers, the generative classifier proposed by Tran \textit{et al.} called DR-GAN learns a disentangled representation~\cite{tran2017disentangled}.

More recent approaches focus on optimizing the embedding distribution. Deng \textit{et al.} proposed using the additive angular margin loss (ArcFace) instead of the commonly used cosine distance loss to improve the discriminative power of the FR model and to stabilize the training process~\cite{deng2019arcface}. Duan \textit{et al.} argued that the distribution of the features plays an important role and therefore proposed using a uniform loss to learn equidistributed representations for their UniformFace FR system~\cite{duan2019uniformface}.

FR systems are vulnerable to presentation attacks, which present an artefact or human characteristic to the biometric (facial) capture subsystem to interfere with the intended policy of the biometric (FR) system\footnote{ISO/IEC CD 30107-1 definition. Accessed at \url{https://www.iso.org/obp/ui/\#iso:std:iso-iec:30107:-1:ed-1:v1:en:term:3.5}}. A photo attack is a presentation attack in which the attacker displays a photograph of the victim to the sensor of the FR system. This photograph can be printed on paper or displayed on a device's screen (\textit{e.g.}, a smartphone, a tablet, or a laptop)~\cite{hernandez2019introduction}. A replay attack is another presentation attack in which a victim's video is played instead of a photograph~\cite{hernandez2019introduction}. A presentation attack detector can be integrated into an FR system to mitigate presentation attacks~\cite{hernandez2019introduction}.

In this work, we used five mainstream high-perform FR systems: Inception-ResNet-v2 networks trained on (1) the CASIA-WebFace database~\cite{yi2014learning} and (2) the MS-Celeb database~\cite{guo2016ms} of de Freitas Pereira \textit{et al.}~\cite{de2018heterogeneous}, (3) the FaceNet network of Sandberg~\cite{sandberg2017facenet}, (4) the DR-GAN of Tran \textit{et al.}~\cite{tran2017disentangled}, and (5) the ArcFace network of Deng \textit{et al.}~\cite{deng2019arcface}. To measure their robustness, besides digital attack, we performed presentation (photo) attacks on them. Please note that there was no presentation attack detector integrated into these FR systems.

\subsection{Wolf attack and master biometric attack}
A ``wolf sample'' is an input sample that can be falsely accepted as a match with multiple user templates (``enrolled subjects") in a biometric recognition system~\cite{une2007wolf}. Wolf samples could be either biometric or non-biometric ones. A wolf sample is used in a wolf attack against a biometric recognition system. An example wolf attack is shown in Fig.~\ref{fig:fr_system}. Wolf attacks were initially used to target fingerprint recognition systems~\cite{ratha2001enhancing}. Its success is theoretically measured by the wolf attack probability (WAP)--the maximum probability of a successful attack with one wolf sample~\cite{une2007wolf}. To mitigate wolf attacks, Inuma \textit{et al.}~\cite{inuma2009theoretical} presented a principle for the construction of secure matching algorithms for any biometric authentication systems by calculating the entropy of the probability distribution of each input value.

A master biometric attack is a wolf attack in which the sample looks like an actual biometric trait. Two example traits are partial fingerprint images~\cite{bontrager2018deepmasterprints} and facial images~\cite{nguyen2020generating}. They are generated by GANs using the LVE algorithm to maximize the false accepting rates (as a result, WAPs are also maximized). A master print attack~\cite{bontrager2018deepmasterprints} targets partial fingerprint recognition systems using small sensors with limited resolution while a master face attack targets FR systems, which require higher resolution images~\cite{nguyen2020generating}. We have extended our previous work~\cite{nguyen2020generating} by using multiple FR systems and databases when running the LVE algorithm. We also simulated presentation attacks using master faces to ascertain their actual threat.

\subsection{Latent variable evolution}
Evolution algorithms are commonly used in artificial intelligence applications to approximate complex, multimodal, and non-differentiable functions since they do not require any assumption about the underlying fitness landscape. The covariance matrix adaptation evolution strategy (CMA-ES) is a powerful strategy designed for non-linear and non-convex functions~\cite{hansen2001completely}. Bontrager \textit{et al.} used CMA-ES with a pre-trained GAN to perform interactive evolutionary computation to improve the quality of generated samples~\cite{bontrager2018deep}. This strategy was used in subsequent work for the LVE algorithm to maximize the WAP of generated partial fingerprint images~\cite{bontrager2018deepmasterprints}. In our previous work~\cite{nguyen2020generating}, we modified the LVE algorithm scoring method so that it could work smoothly with high-resolution facial images generated by StyleGAN~\cite{karras2019style}.

Given $n$ random initial vectors $\mathcal{Z} = \{\mathbf{z}_1, \mathbf{z}_2, ..., \mathbf{z}_n\}$, a generation model $\mathcal{G}$, a scoring function $\mathcal{F}$, and $m$ enrolled temples $\mathcal{T} = \{\mathbf{t}_1, \mathbf{t}_2, ...,\mathbf{t}_m\}$. The LVE algorithm runs in a loop in which at first, $n$ samples are generated by $\mathcal{G}$ using $\mathcal{Z}$. Each sample is then matched with $m$ templates in $\mathcal{T}$ to obtain a mean score $s$. An evolution algorithm (\textit{e.g.}, CMA-ES) takes the set of the mean scores $s$ to evolve $n$ new latent vectors $\mathcal{Z'}$ for the next loop.

We have now added one more surrogate FR system and database to the LVE algorithm to better approximate the target FR system and database so that the generated master faces could have better generalizability.

\section{Deep master faces}
\label{sec:deep_master_faces}
\subsection{Existence of master faces}
\begin{figure}[t]
	\centering
	\includegraphics[width=85mm]{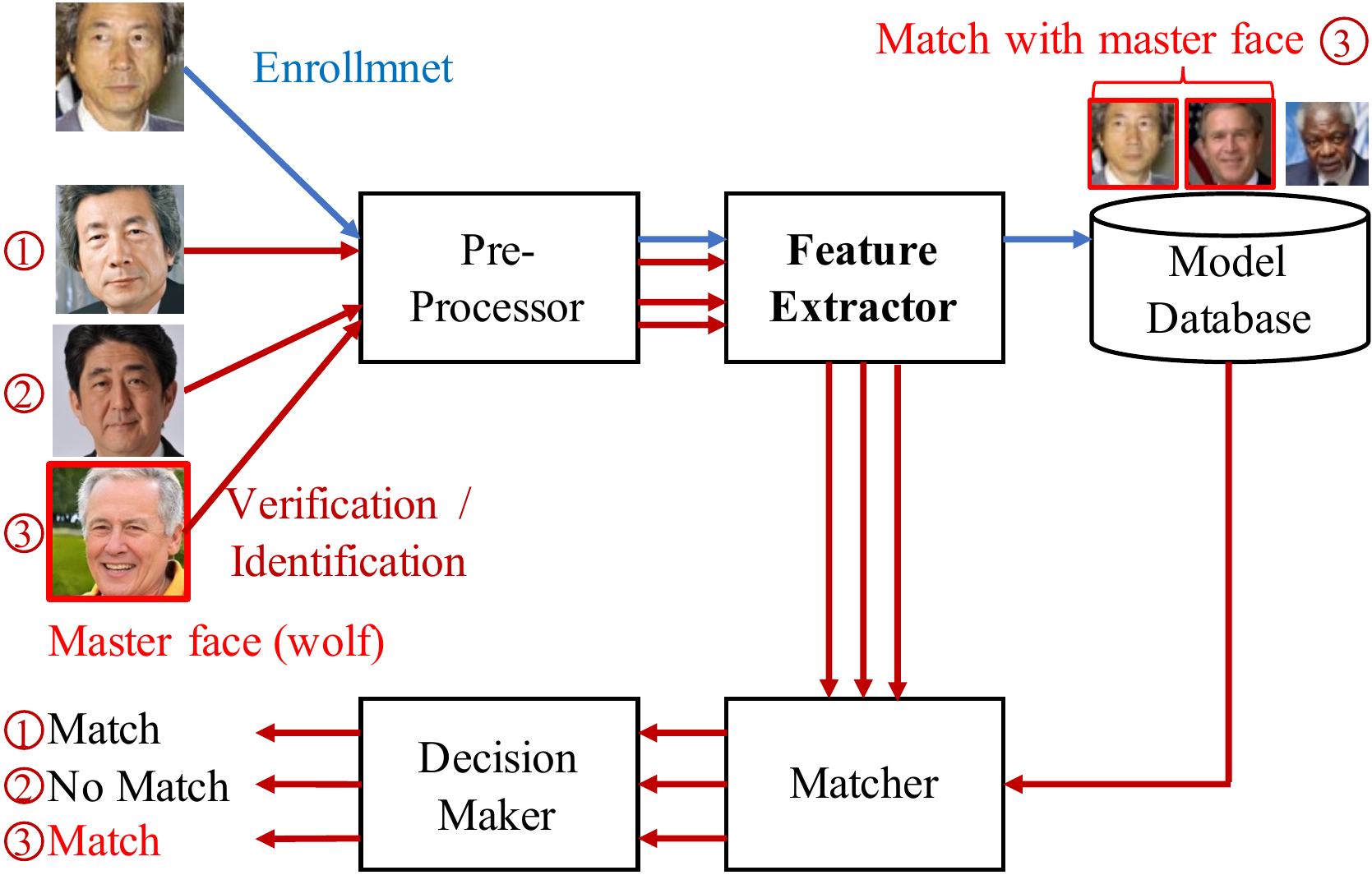}
	\caption{Operation of typical FR system. There are two phases: enrollment (blue path) and verification/identification (red path). The master face (face 3) was falsely matched with the enrolled subject. Best viewed in color.}
	\label{fig:fr_system}
\end{figure}

Before describing the proposed master face generation algorithm, we briefly explain why the master face exists. For a typical FR system (or biometric recognition systems in general), there are four phases (Fig.~\ref{fig:fr_system}): pre-processing the input, extracting its features, matching them with those of the enrolled subject(s) in the model database, and making a decision. The feature extractor plays the role of a mapping function. It maps the facial image domain to the identity domain. The objective when training the feature extractor is to optimize the mapping function so that the mappings of the same-identity faces are close together in the identity space and vice versa. Since this is an optimization problem, the solution is simply an approximation. Furthermore, there is no guarantee that the mapping function will work well on new data due to the possibility of lack of generalizability.

\begin{figure}[t]
	\centering
	\includegraphics[width=85mm]{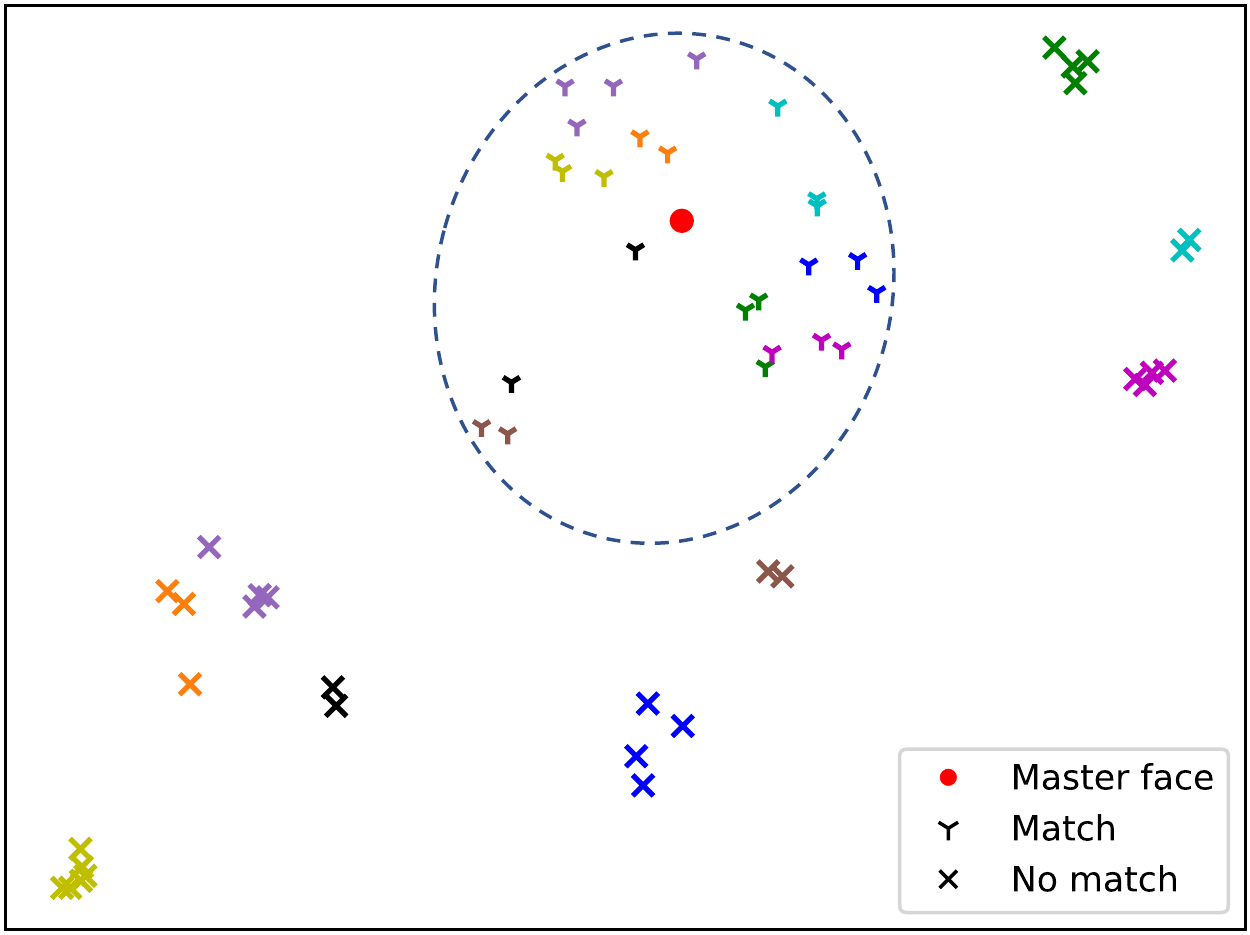}
	\caption{UMAP visualization of identity space containing embeddings of a master face and of ``match" and ``no-match" faces of 18 enrolled subjects. For each cluster (match or no match), symbols with the same color correspond to the same subject. Best viewed in color.}
	\label{fig:embeddings}
\end{figure}

Master faces may exist because the identity (embedding) space used by FR systems is not uniformly distributed, resulting in dense areas in this space. If we generate an identity corresponding to a point in a dense area, it may falsely match several nearby faces in the identity space. The LVE algorithm aims to find such a position in a dense area in the identity space after several evolutions. To intuitively and empirically show this, we visualize the identity space and one of the master face generated in this paper using uniform manifold approximation and projection (UMAP)~\cite{mcinnes2018umap} in Fig.~\ref{fig:embeddings}.  The master face generated by our algorithm described in the next section is at such a position (red dot) surrounded by many embeddings. All faces from these surrounding embeddings are falsely matched with the master face by the FR system. The no-match embeddings are scattered far from the master face and lay in less dense areas.  We explain how to generate such master faces in the next section.

\subsection{Latent variable evolution with multiple databases and/or face recognition systems}
\begin{figure}[t]
	\centering
	\includegraphics[width=85mm]{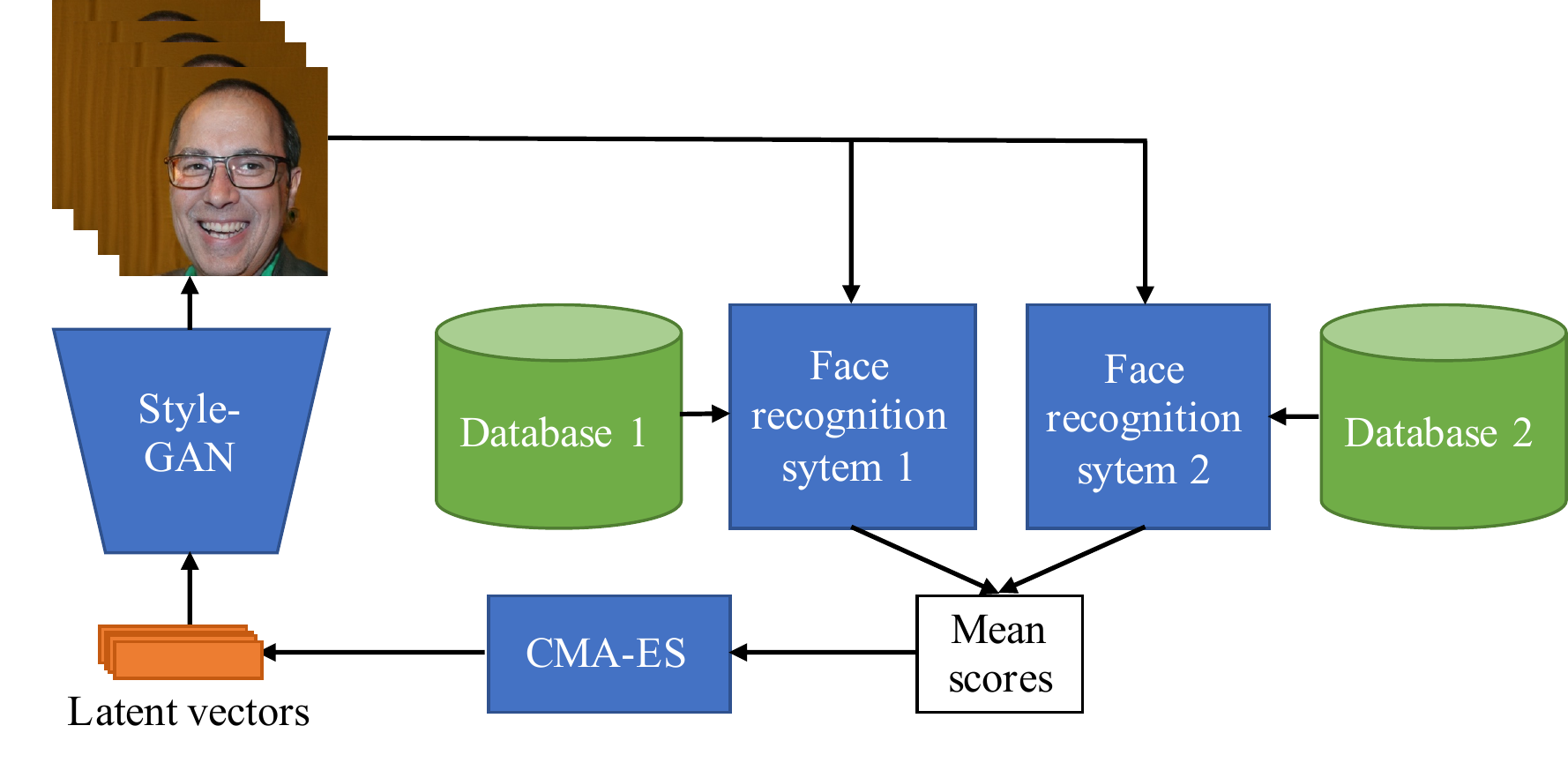}
	\caption{Overview of extended latent variable evolution algorithm. Latent vectors are fed into StyleGAN~\cite{karras2019style} to generate facial images. One or more surrogate FR system(s) then calculates mean score for each image on the basis of the subjects in one or more database(s). For example, for the \textit{combination 3} setting described in Table~\ref{tab:combined_detail}, \textbf{database 1} is LFW - Fold 1, \textbf{database 2} is MOBIO, \textbf{FR system 1} is Inception-ResNet-v2 network (trained on MS-Celeb database), and \textbf{FR system 2} is DR-GAN network. The CMA-ES~\cite{hansen2001completely} algorithm uses these scores to generate new latent vectors.}
	\label{fig:lve}
\end{figure}

We extended our previous work by adding one more database and/or FR system to generate master faces, which requires support from the LVE algorithm. The extended LVE algorithm is illustrated in Fig.~\ref{fig:lve} and is formalized in Algorithm~\ref{alg:lve}. First, $m$ latent vectors $\{\mathbf{z_1},...,\mathbf{z_m}\}$ are initialized randomly. Then, they are fed into a pre-trained StyleGAN network to generate $m$ faces. Two face matching functions, $FaceMatching^{(1)}(\cdot,\cdot)$ and $FaceMatching^{(2)}(\cdot,\cdot)$ (corresponding to two FR systems), calculate the similarity between the generated faces and all subject faces in databases $E_j^{(1)}$ and $E_j^{(2)}$, respectively. Two $m$-dimension mean score vectors, $\mathbf{s^{(1)}}$ and $\mathbf{s^{(2)}}$, are obtained from the results of $FaceMatching^{(1)}(\cdot,\cdot)$ and $FaceMatching^{(2)}(\cdot,\cdot)$. The mean $\mathbf{s}$ of these two vectors is used to select the best local master face $F_b$ among the $m$ generated faces. Finally, $\mathbf{s}$ is fed into the CMA-ES algorithm to generate new latent vectors $\{\mathbf{z_1},...,\mathbf{z_m}\}$. This process is repeated $n$ times. The final (global) best master face is chosen from among the $n$ best master faces $\mathcal{F}$ obtained in the $n$ iterations.

\begin{algorithm}
	\caption{Latent variable evolution.}
	\label{alg:lve}
	\begin{algorithmic}
		\State $m \gets 22$
		\Comment{Population size}
		
		\Procedure{RunLVE}{$m, n$}
		\State $\mathcal{F} = \{\}$
		\Comment{Master face set}
		\State $\mathcal{S} = \{\}$
		\Comment{and corresponding score set}
		
		\State $\mathcal{Z} = \{\mathbf{z_1} \gets rand(),...,\mathbf{z_m} \gets rand()\}$
		\Comment{Initialize} 
		
		\For {$n$ iterations}
		\Comment{Run LVE algorithm $n$ times}
		\State $F \gets StyleGAN(\mathcal{Z}$)
		\Comment{Generate $m$ faces $F$}
		
		\State $\mathbf{s^{(1)}} \gets 0, \mathbf{s^{(2)}} \gets 0$
		\Comment{Initialize scores $\mathbf{s^{(1)}}, \mathbf{s^{(2)}} \in \R^{m}$}
		
		\For {face $F_i$ in faces $\mathbf{F}$}
		\For{face $E_j^{(1)}$ in data $\mathbf{E^{(1)}}$}
		\State $\mathbf{s^{(1)}} \gets \mathbf{s^{(1)}} + FaceMatching^{(1)}$($F_i, E_j^{(1)}$)
		\EndFor
		
		\State $\mathbf{s^{(1)}} \gets \frac{\mathbf{s^{(1)}}}{|\mathbf{E^{(1)}}|}$
		\Comment{Mean scores of 1st system}
		
		\For{face $E_j^{(2)}$ in data $\mathbf{E^{(2)}}$}
		\State $\mathbf{s^{(2)}} \gets \mathbf{s^{(2)}} + FaceMatching^{(2)}$($F_i, E_j^{(2)}$)
		\EndFor
		
		\State $\mathbf{s^{(2)}} \gets \frac{\mathbf{s^{(2)}}}{|\mathbf{E^{(2)}}|}$
		\Comment{Mean scores of 2nd system}
		
		\State $\mathbf{s} = \frac{\mathbf{s^{(1)}} + \mathbf{s^{(2)}}}{2}$
		\Comment{Mean scores of both systems}
		
		\State $F_b, s_b \gets GetBestFace(\mathbf{F}, \mathbf{s}$)
		
		\State $\mathcal{F} \gets  \mathcal{F} \cup \{F_b\}$
		\Comment{Append best master face}
		\State $\mathcal{S} \gets  \mathcal{S} \cup \{s_b\}$
		\Comment{and its corresponding score}
		
		\State $\mathcal{Z} \gets$ CMA\_ES($\mathbf{s}$)
		\EndFor
		\EndFor
		
		\State \textbf{return} $\mathcal{F}, \mathcal{S}$
		\EndProcedure
		
		\State $F_b, s_b \gets GetBestFace(\mathcal{F}, \mathcal{S})$
		\Comment{Final (best) master face}
	\end{algorithmic}
\end{algorithm}

\section{Generating Master Faces}
\label{sec:generating}

To evaluate the risks and threats of the master face attacks, we designed several settings for the LVE algorithm and several attack scenarios so that they can cover white-box, gray-box, and black-box attacks. For white-box attacks, both the architecture of the target FR system and its training database are known while for gray-box attacks, only one of them is known. For black-box attacks, there is no information about the target FR system. Attackers may use more than one FR system for the LVE algorithm to increase the probability of their attack being a white-box or gray-box attack. They can also use more than one database for the LVE algorithm to better approximate the distribution of the model database of the target FR system.

The structure of this section is organized as follows: We first briefly describe the information of the FR systems and the databases we used in our experiments. Then, we describe our generation of master faces using several combinations of single and multiple FR systems with single and multiple facial databases when running the LVE algorithm. After that, we analyze the generation processes and the generated master faces as well as provide some explanation of their properties. Finally, we evaluate the false-matching performance of the generated master faces for several scenarios, including black-box, gray-box, and white-box attacks.

\subsection{Experiment Materials}

\subsubsection{Face recognition systems}
We used five mainstream publicly available high-performance FR systems in our experiments:
\begin{itemize}
\item Inception-ResNet-v2 based FR systems: one trained on the CASIA-WebFace database~\cite{yi2014learning} and one trained on the MS-Celeb database~\cite{guo2016ms} by de Freitas Pereira \textit{et al.}~\cite{de2018heterogeneous}.

\item Open-source version of FaceNet~\cite{schroff2015facenet} implemented and trained on the MS-Celeb database~\cite{guo2016ms} by Sandberg~\cite{sandberg2017facenet}.

\item DR-GAN~\cite{tran2017disentangled} trained on a combination of the Multi-PIE database~\cite{gross2010multi} and the CASIA-WebFace database~\cite{yi2014learning}.

\item ArcFace~\cite{deng2019arcface} trained on the MS-Celeb database~\cite{guo2016ms}.

\end{itemize}

We used two Inception-ResNet-v2 based FR systems and DR-GAN for generating master faces and all of the above FR systems for evaluating master face attacks\footnote{A benchmark for some of them can be found at \url{https://www.idiap.ch/software/bob/docs/bob/bob.bio.face_ongoing/v1.0.4/leaderboard.html}}. They were all pre-trained and obtained from the Bob toolbox~\cite{bob2017}.

\subsubsection{Databases}

\begin{table}[t]
\centering
    \caption{Details of all databases used in our experiments.}
    \label{tab:databases}
    \begin{tabular}{l|l|c|c}
    \hline
    \multicolumn{1}{c|}{\textbf{Database}} & \multicolumn{1}{c|}{\textbf{Year}} & \textbf{No. of images} & \textbf{Resolution} \\
    \hline
    Flickr-Faces-HQ~\cite{karras2019style} & 2019 & 70,000 & 1024 $\times$ 1024 \\
    CASIA-WebFace~\cite{yi2014learning} & 2014 & 494,414 & 256 $\times$ 256 \\
    MS-Celeb~\cite{guo2016ms} & 2016 & 10,490,534 & Up to 300 $\times$ 300 \\
    Multi-PIE~\cite{gross2010multi} & 2009 & 755,370 & 3072 $\times$ 2048 \\
    LFW~\cite{LFWTechUpdate} & 2007 & 13,233 & Various \\
    MOBIO~\cite{mccool2012bi} & 2012 & 30,326 & Various \\
    IJB-A~\cite{klare2015pushing} & 2015 & 5,712 & Various \\ \hline
    \end{tabular}
\end{table}

\begin{figure}[th!]
	\centering
	\includegraphics[width=70mm]{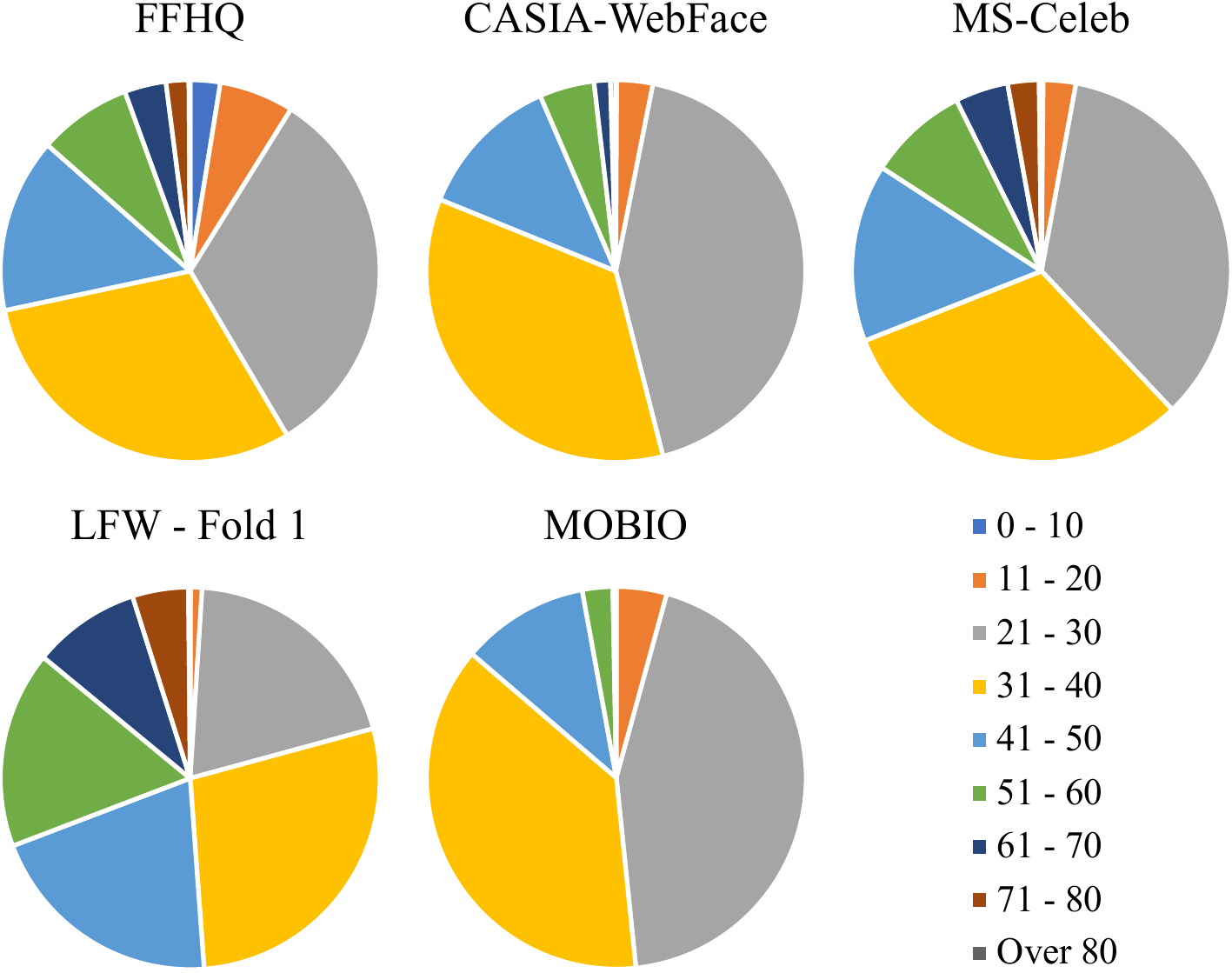}
	\caption{Estimated age distribution of five databases used for training StyleGAN, FR systems, and generation of master faces. Best viewed in color.}
	\label{fig:age}
\end{figure}

\begin{figure}[th!]
	\centering
	\includegraphics[width=70mm]{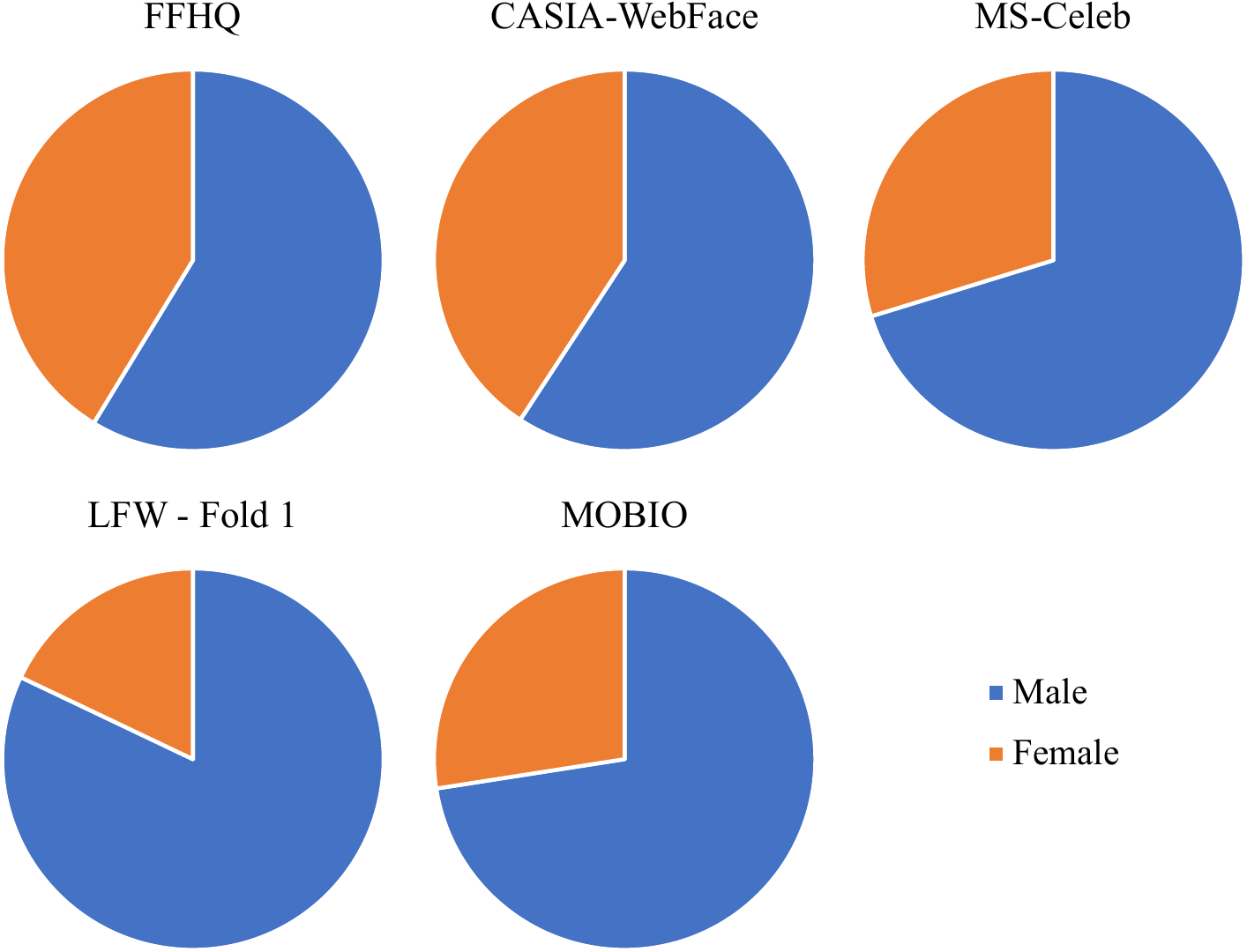}
	\caption{Estimated gender distribution of five databases used for training StyleGAN, FR systems, and generation of master faces. Best viewed in color.}
	\label{fig:gender}
\end{figure}

Seven databases were used for four different purposes:
\begin{itemize}
\item \textit{Training StyleGAN:} Flickr-Faces-HQ (FFHQ) database~\cite{karras2019style}.
\item \textit{Training FR systems:} CASIA-WebFace~\cite{yi2014learning}, MS-Celeb~\cite{guo2016ms}, and Multi-PIE~\cite{gross2010multi}.
\item \textit{Running LVE algorithm:} The training set of the Labeled Faces in the Wild (LFW) - Fold 1 database~\cite{LFWTechUpdate} aligned by funneling~\cite{Huang2007a} and both the male and female components of the training set of the mobile biometry (MOBIO) database~\cite{mccool2012bi}.
\item \textit{Evaluating master faces:} The corresponding development (dev) and evaluation (eval) sets of the LFW database~\cite{LFWTechUpdate} and the MOBIO database~\cite{mccool2012bi} plus the dev set of the IARPA Janus Benchmark A (IJB-A) database~\cite{klare2015pushing}\footnote{there is no eval set for the IJB-A database}. The dev sets were used for threshold selection for the FR systems (which was based on the calculated equal error rates, EERs).
\end{itemize}

Details of all databases are shown in Table~\ref{tab:databases}. There are no overlapping subjects between the databases used for training StyleGAN, training the FR systems, and running the LVE algorithm. This point is important to demonstrate that the LVE algorithm can work well with mutually exclusive databases used by its components.

We used the InsightFace library\footnote{\url{https://github.com/deepinsight/insightface}} to estimate the age and gender distributions of the databases used for training StyleGAN, FR systems, and generation of master faces. For the MOBIO one, we used its annotated gender information instead. We ignored the Multi-PIE database since it only contributes as an additional part of the training database of the DR-GAN FR system. The estimated distributions are shown in Fig.~\ref{fig:age} and Fig.~\ref{fig:gender} respectively. The ages are dominantly 21 to 40, especially in the CASIA-WebFace, MS-Celeb, and MOBIO ones. The LFW - Fold 1 database is more balanced with a larger proportion of 41 to 60 ages. There are tiny numbers of child faces in all databases except for the MOBIO one, which has none. For gender, there are more male than female faces in all databases. The LFW - Fold 1 and MOBIO databases are the most unbalanced with less than 25\% female faces. This may cause bias in the FR systems as well as affect the properties of the generated master faces, as explained in the following section.

\begin{table*}[th!]
    \centering
    \caption{Settings for running LVE algorithm. ``Single" means using only one database and one FR system. ``Combination" means using more than one database and/or FR system. For each FR system, we show both its network architecture (top row) and its training database (bottom row).}
    \label{tab:settings}
    \begin{tabular}{c|c|cc|cc}
        \hline
        \textbf{No.} & \textbf{Setting} & \textbf{Database 1} & \textbf{\begin{tabular}[c]{@{}c@{}}FR System 1\\ (FR Training DB)\end{tabular}} & \textbf{Database 2} & \textbf{\begin{tabular}[c]{@{}c@{}}FR System 2\\ (FR Training DB)\end{tabular}} \\ \hline
        1 & \textit{Single 1} & LFW - Fold 1 & \begin{tabular}[c]{@{}c@{}}Inception-ResNet-v2\\ (CASIA-WebFace)\end{tabular} & & \\ \hdashline
        2 & \textit{Single 2} & LFW - Fold 1 & \begin{tabular}[c]{@{}c@{}}DR-GAN\\ (CASIA-WebFace \& Multi-PIE)\end{tabular} & & \\ \hdashline
        3 & \textit{Single 3} & MOBIO & \begin{tabular}[c]{@{}c@{}}Inception-ResNet-v2\\ (MS-Celeb)\end{tabular} & & \\ \hdashline
        4 & \textit{Single 4} & LFW - Fold 1 & \begin{tabular}[c]{@{}c@{}}Inception-ResNet-v2\\ (MS-Celeb)\end{tabular} & & \\ \hdashline
        5 & \textit{Single 5} & MOBIO & \begin{tabular}[c]{@{}c@{}}DR-GAN\\ (CASIA-WebFace \& Multi-PIE)\end{tabular} & & \\ \hline
        6 & \begin{tabular}[c]{@{}c@{}}\textit{Combination 1}\\ (No. 1 \& 2)\end{tabular} & LFW - Fold 1 & \begin{tabular}[c]{@{}c@{}}Inception-ResNet-v2\\ (CASIA-WebFace)\end{tabular} & LFW - Fold 1 & \begin{tabular}[c]{@{}c@{}}DR-GAN\\ (CASIA-WebFace \& Multi-PIE)\end{tabular} \\ \hdashline
        7 & \begin{tabular}[c]{@{}c@{}}\textit{Combination 2}\\ (No. 1 \& 3)\end{tabular} & LFW - Fold 1 & \begin{tabular}[c]{@{}c@{}}Inception-ResNet-v2\\ (CASIA-WebFace)\end{tabular} & MOBIO & \begin{tabular}[c]{@{}c@{}}Inception-ResNet-v2\\ (MS-Celeb)\end{tabular} \\ \hdashline
        8 & \begin{tabular}[c]{@{}c@{}}\textit{Combination 3}\\ (No. 4 \& 5)\end{tabular} & LFW - Fold 1 & \begin{tabular}[c]{@{}c@{}}Inception-ResNet-v2\\ (MS-Celeb)\end{tabular} & MOBIO & \begin{tabular}[c]{@{}c@{}}DR-GAN\\ (CASIA-WebFace \& Multi-PIE)\end{tabular} \\ \hline
    \end{tabular}
\end{table*}

\begin{table}[th!]
    \centering
    \caption{Comparison between three combination settings for LVE algorithm. For FR systems, we compared their architectures and their training databases.}
    \label{tab:combined_detail}
    \begin{tabular}{c|c|cc}
        \hline
        \multirow{2}{*}{\textbf{Setting}} & \multirow{2}{*}{\textbf{\begin{tabular}[c]{@{}c@{}}Database 1 vs.\\ Database 2\end{tabular}}} & \multicolumn{2}{c}{\textbf{\begin{tabular}[c]{@{}c@{}}FR System 1 vs.\\ FR System 2\end{tabular}}} \\ \cline{3-4} 
        & & \textbf{Architectures} & \textbf{Training DBs} \\ \hline
        \textit{Combination 1} & Same & Different & Similar \\
        \textit{Combination 2} & Different & Same & Different \\
        \textit{Combination 3} & Different & Different & Different \\ \hline
    \end{tabular}
\end{table}

\subsection{Latent Variable Evolution Configurations}
\label{sec:lve_config}

Since there are many FR systems and databases, evaluating all possible combinations is impossible with the available computation and time resources. We thus selected a subset with the aim of covering a range as broad as possible. We defined eight settings (Table~\ref{tab:settings}) for the LVE algorithm using three FR systems (two version of Inception-ResNet-v2, one trained on the CASIA-WebFace database and one trained on the MS-Celeb one, and DR-GAN) and two databases (LFW - Fold 1 and MOBIO). There are five settings in which one FR system and one database are used (\textit{single 1} to \textit{single 5}) and three settings in which more than one FR system and/or database is used (\textit{combination 1}, \textit{combination 2}, and \textit{combination 3}).
Each combination setting combined two single settings and was selected on the basis of their reasonable coverage of cases. The main differences of the three combination settings are highlighted in Table~\ref{tab:combined_detail}. In the \textit{combination 1} setting, only one database was used with the LVE algorithm, and the databases used for training the FR systems were similar. In the \textit{combination 2} setting, two databases were used with the LVE algorithm, and two FR systems with the same architecture but trained on different databases were used. In the \textit{combination 3} setting, two databases and two FR systems without anything in common were used with the LVE algorithm. We ran 1000 iterations of the LVE algorithm for each of the eight settings.

\begin{figure}[t]
	\centering
	\includegraphics[width=89mm]{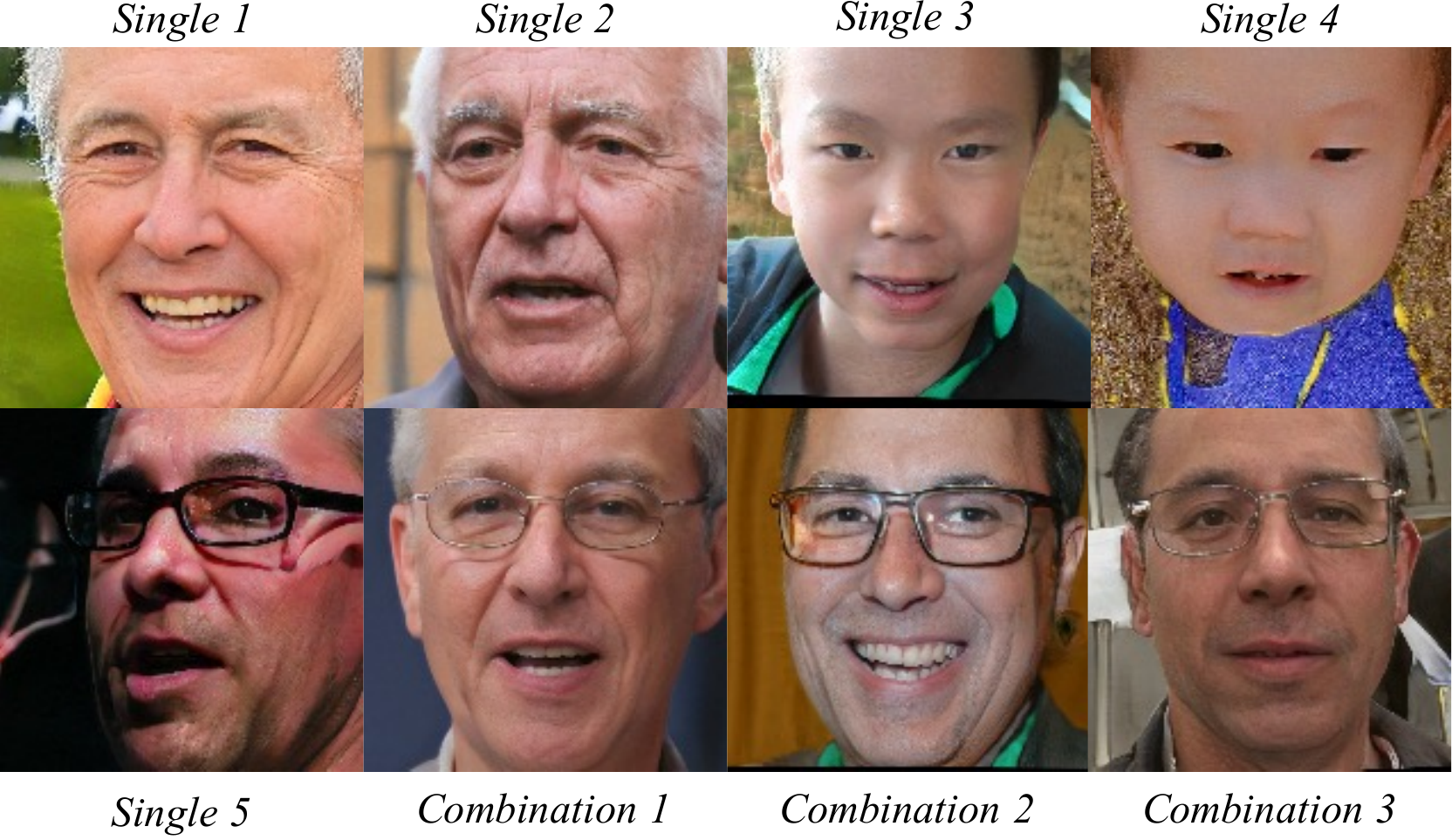}
	\caption{Master faces generated using eight settings specified in Table~\ref{tab:settings}.}
	\label{fig:all_masterfaces}
\end{figure}

\begin{figure}[th!]
	\centering
	\includegraphics[width=89mm]{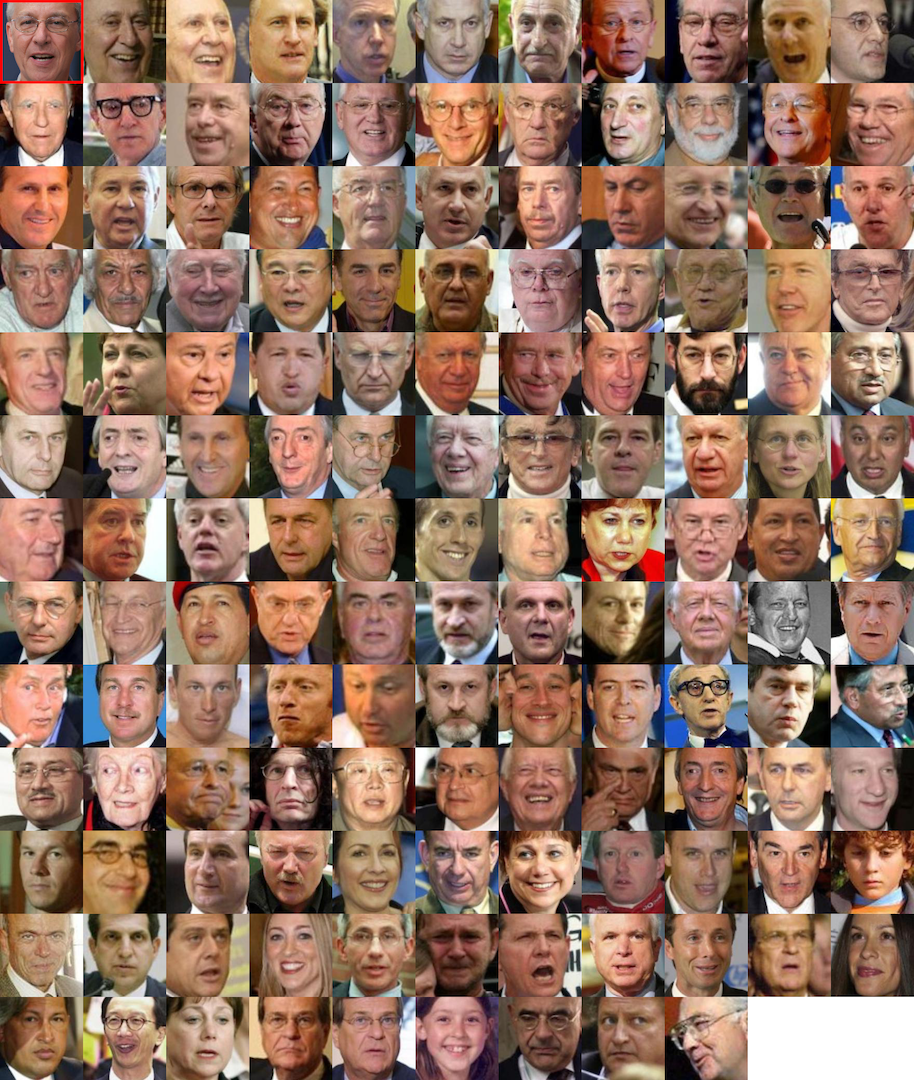}
	\caption{Master face (top left) generated using \textit{combination 1} setting and all matched faces from eval set of the LFW - Fold 1 database~\cite{LFWTechUpdate} sorted from closest to farthest match. Inception-ResNet-v2 based FR system~\cite{de2018heterogeneous} was used in this case.}
	\label{fig:matches}
\end{figure}

The generated master faces corresponding to the eight settings described earlier are shown in Fig.~\ref{fig:all_masterfaces}. All of them are male faces. One-fourth are child faces, generated using only the Inception-ResNet-v2 based FR system trained on the MS-Celeb database. Half are senior faces, generated using only the Inception-ResNet-v2 based FR system trained on the CASIA-WebFace database, or only the DR-GAN FR system trained on the combination of the CASIA-WebFace and Multi-PIE databases, or a combination of these two FR systems. The rest (one-fourth) are middle-aged faces, generated using the combinations of the two FR systems in the previous two cases (one in each case).

\subsection{Master Face Analysis}
\label{sec:analysis}

The master face generated using the \textit{combination 1} setting and its matched faces by using the Inception-ResNet-v2 based FR system~\cite{de2018heterogeneous} on the eval set of the LFW - Fold 1 database~\cite{LFWTechUpdate} are shown in Fig.~\ref{fig:matches}. This master face matched those of persons of both genders, of multiple races (Caucasian, Black, and Asian), and of multiple ages (from children to seniors). In many cases, the facial angles and lighting conditions differed from those of the master face. The subjects are both wearing and not wearing glasses (normal glasses or sunglasses). A typical master face can match about 10 to 50 identities.

\begin{figure}[th!]
	\centering
	\includegraphics[width=50mm]{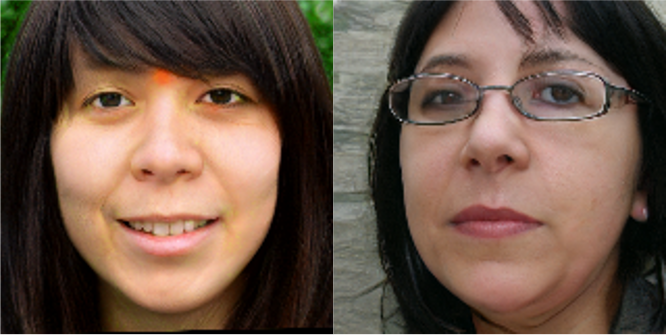}
	\caption{Two female master faces generated using only female part of MOBIO database and Inception-ResNet-v2 based FR system (MS-Celeb version) and DR-GAN FR system, respectively.}
	\label{fig:women}
\end{figure}

\begin{figure*}[th!]
	\centering
	\includegraphics[width=145mm]{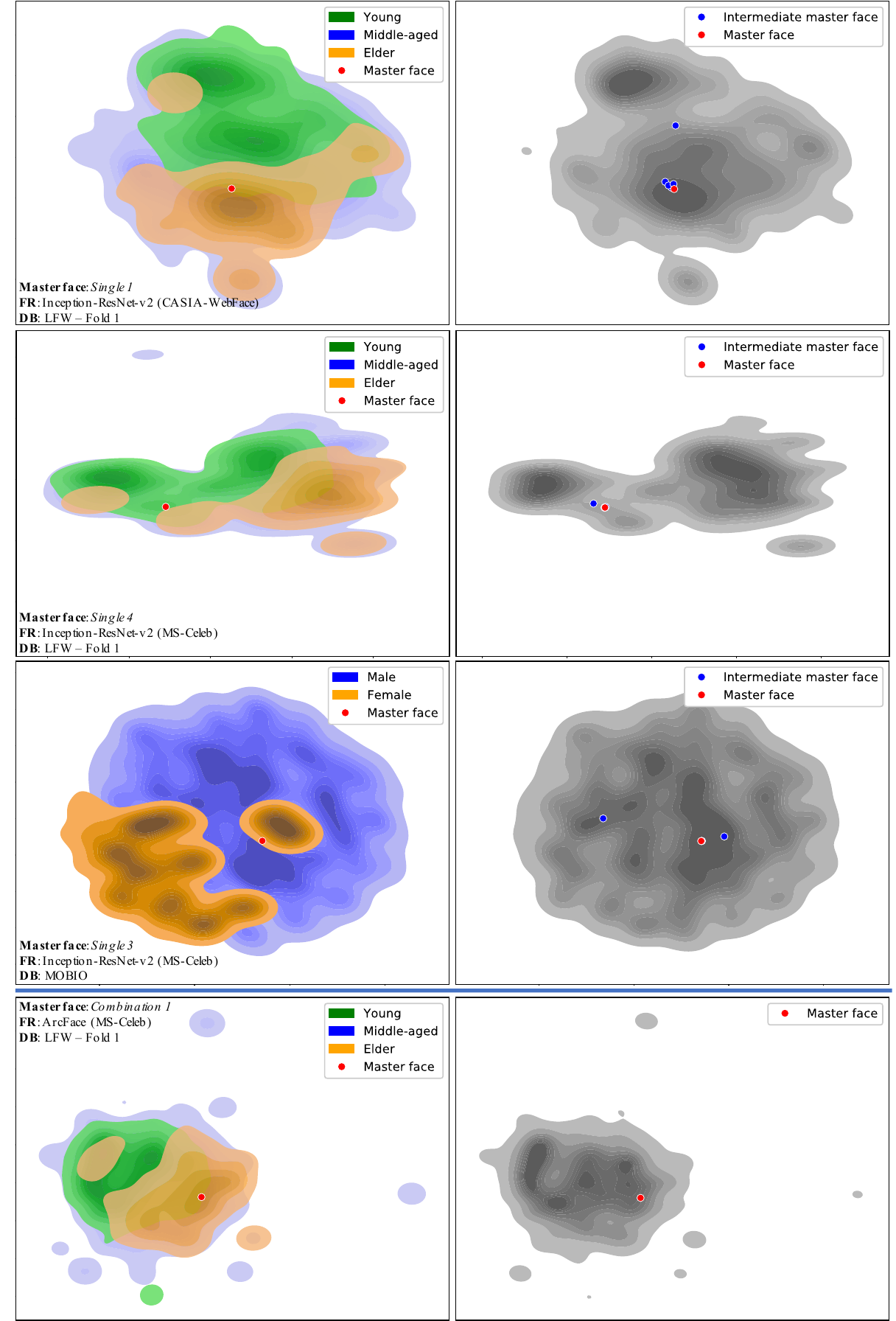}
	\caption{Estimated densities of ages (row 1, 2, and 4) and genders (row 3) of the embedded faces extracted by the Inception-ResNet-v2 based FR systems trained on the CASIA-WebFace database (row 1) and the MS-Celeb database (row 2 and 3), and by the ArcFace FR system (row 4). Plots on the left-hand side show estimated densities per class, while those on the right-hand side show estimated densities of all embeddings. Two Inception-ResNet-v2 based FR systems were used in both attacker and defender sides, while the ArcFace one was used only on the defender side. We also included the embeddings of five intermediate master faces generated during running the LVE algorithm (blue dots) and the optimized master face (red dot). These embeddings were extracted from the entire training sets of the LFW - Fold 1  database (row 1, 2, and 4) and of the MOBIO database (row 3). Corresponding LVE settings (see Table~\ref{tab:settings} for more detail) used to generate master faces are shown in the left figures, along with information about the target FR systems (denoted as FRs) and databases (denoted as DBs). Best viewed in color.}
	\label{fig:density}
\end{figure*}

To better understand these results, we performed the UMAP dimension reduction algorithm on the embedding spaces of some FR systems, then applied a kernel density estimation method on the reduced spaces to form the density maps. We did that from both age and gender perspectives. We chose two Inception-ResNet-v2 based FR systems (CASIA-WebFace version and MS-Celeb version) and the ArcFace FR system to perform the embedding space density estimation. Among them, the two Inception-ResNet-v2 FR systems were used on both the attacker side and the defender side, while the ArcFace FR system was only used on the defender side. The estimated densities are shown in Fig.~\ref{fig:density}. We also included the positions of intermediate master faces' and the optimized master faces' embeddings in the plots.

From the age perspective, young faces (less than 30 years old) are separated from the senior faces (more than 60 years old) while the remaining faces (30 to 60 years old) are scattered throughout both the young and senior faces.
From the gender perspective, the male and female faces are somewhat separated. To maximize the false matches, the LVE algorithm placed the master face in a dense area near the border of a cluster, which increased the probability of matching diverse faces. Since there are more male than female faces in all the databases, the probability of placement in a dense area in the male cluster was higher than that of placement in the female one. However, since they were only somewhat separated, the master faces could match both male and female faces (with more male face matches, as shown in Fig.~\ref{fig:matches}). For age, the selected dense area could be in a young cluster, a middle-aged cluster, or an elder cluster. Since the training data for the FR systems was unbalanced in terms of age with only a few samples for young and senior faces, these systems may not accurately recognize those faces. The CASIA version of the Inception-ResNet-v2 based FR system may perform poorly on senior male faces, resulting in the generation of senior male master faces. Interestingly, the master face generated using the \textit{combination 1} also lays at the centroid of the ArcFace FR system, which is only used on the defender side. For this case, dense areas also exist even if we use the angular margin loss in training. On the other hand, the MS-Celeb version of the Inception-ResNet-v2 based FR system performs poorly on young male faces, resulting in the generation of boy master faces. For the \textit{combination 2} (not fully shown in Fig.~\ref{fig:density} due to limited space), we observed that the 30 to 60 year-old faces were scattered in the embedding spaces of both of these FR systems, it seems that an ``average" middle-aged face is the optimal solution according to the proposed LVE algorithm. To further verify the effects of the clusters on the properties of the master faces, we generated two master faces using only the female part of the MOBIO database and the Inception-ResNet-v2 FR system (MS-Celeb version) and the DR-GAN FR system. Both master faces are female, as shown in Fig~\ref{fig:women}.

\subsection{False Matching Rate Analysis}
\label{sec:fmr_analysis}

\begin{figure*}[th!]
	\centering
	\includegraphics[width=171mm]{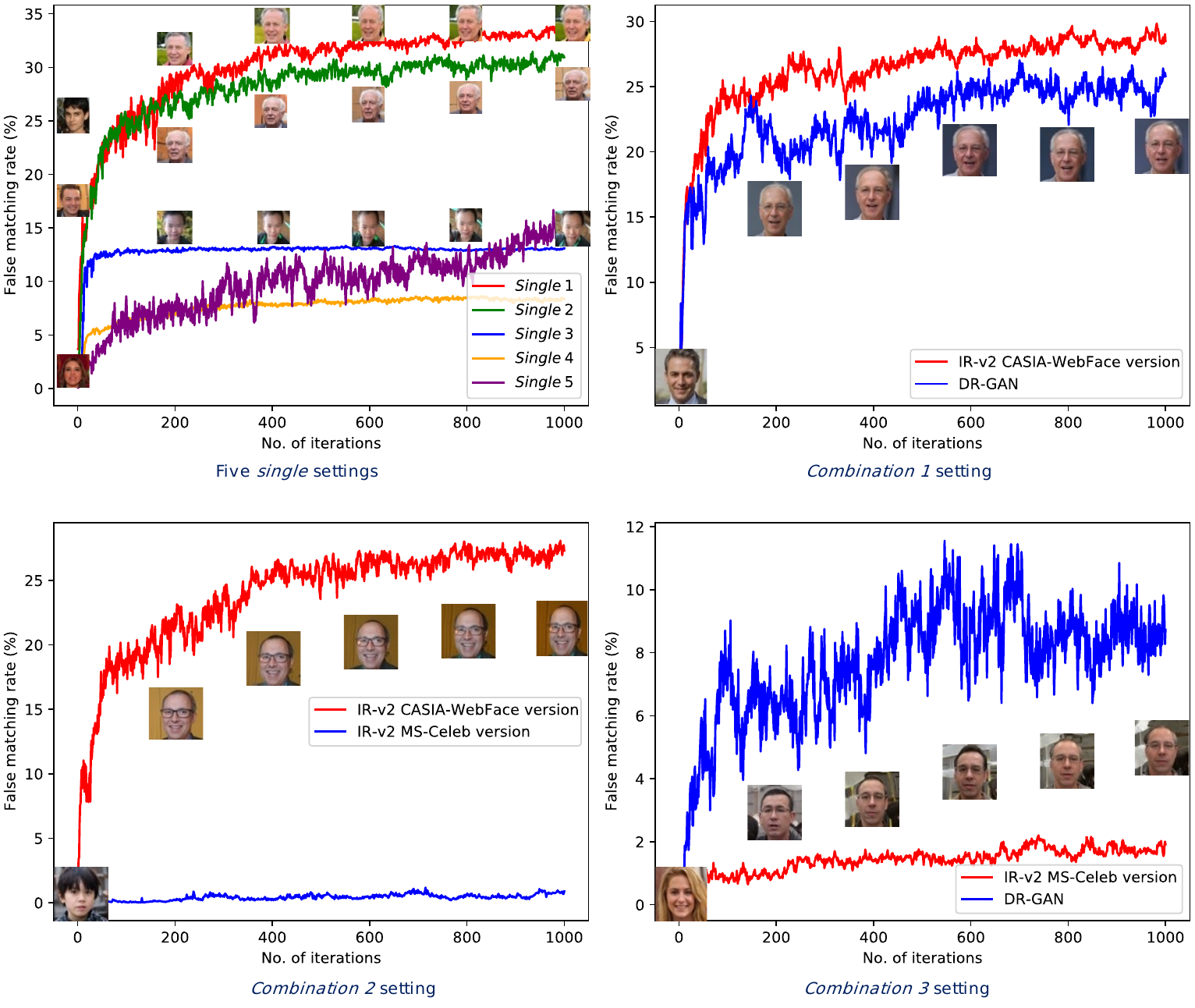}
	\caption{FMRs of each FR system when running the LVE algorithm using five \textit{single} settings and three \textit{combination} settings. There are two Inception-ResNet-v2 (IR-v2) FR systems, one trained on  CASIA-WebFace database and one trained on MS-Celeb database. We included intermediate master faces generated using three \textit{single} settings 1, 2, and 3, and three combination settings. Best viewed in color.}
	\label{fig:fmr}
\end{figure*}

Next, we evaluate the performances of attacks using the master faces. If the generated master face matches more enrolled users, false matching rates (FMRs) of the FR systems become higher. Hence we compared the FMRs between two tests:
\begin{itemize}
\item \textit{Normal test:} One side of the test pairs included either genuine or zero-effort imposter faces defined by the test protocols of the current database.
\item \textit{Master face test:} The master face was paired with all the enrolled faces.
\end{itemize}

First, we show how the FMRs measured on the master face set changed during the LVE optimization in Fig.~\ref{fig:fmr}.  As we can see from the figure, the FMRs become higher in six out of the eight settings. For the two remaining settings (\textit{combination 2} and \textit{3}), the FMR of one of their component FR systems also becomes higher while this of the other component FR system remains almost zero. In these two cases, two different databases were used with the LVE algorithm, and the algorithm tried to maximize the similarities between the master face and all faces in the database 1 as calculated by the component FR system 1 as well as to maximize the similarities between the master face and all faces in the database 2 as calculated by the component FR system 2. This task is difficult, even if the two FR systems share the same architecture, as they do with the \textit{combination 2} setting.
Since the LFW and MOBIO databases have different distributions, finding a master face that matches many subjects in both of them is challenging. The LVE algorithm could focus only on one database (the LFW database) and ignore the other (the MOBIO database, which has higher variability in terms of pose and illumination conditions than the LFW database). Moreover, the Inception-ResNet-v2 based FR system trained on the MS-Celeb database was harder to fool when running it with the LVE algorithm compared with its CASIA-WebFace version. In contrast, although two FR systems were used in the \textit{combination 1} setting, they shared the same database, so the algorithm was able to fool both of them.

Two rules for designing settings for the LVE algorithm can be inferred from these results:
\begin{itemize}
\item Using more than one database for running the LVE algorithm is difficult. The algorithm may prioritize the database that is less challenging.
\item Using more than two FR systems is OK. They can have the same or different architectures, trained on similar or different databases.
\end{itemize}

\begin{table*}[th!]
\centering
\caption{FMRs of normal tests and the corresponding master face tests using master faces generated using five \textit{single} settings and three \textit{combination} settings. For each FR system, we show both its network architecture (top row) and its training database (bottom row). Within each cell, numbers on the upper part, the lower part, the left, and the right are the FMRs of the normal tests and the master face tests, from the development sets and the evaluation sets of the target databases, respectively. Gray cells indicate the surrogate database(s) used by attackers when running the LVE algorithm and the target database(s) are different, while gray cells imply that they are the same. Numbers in blue indicate that the surrogate FR system(s) and the target FR system(s) are identical in both architecture(s) and training database(s). Numbers in bold indicate success master face attacks. Best viewed in color.}
\label{tab:fmrs}
\adjustbox{max width=\textwidth}{

\begin{tabular}{l|l|rr;{1pt/1pt}rr;{1pt/1pt}rr;{1pt/1pt}rr;{1pt/1pt}rr|rr;{1pt/1pt}rr;{1pt/1pt}rr} 
\hline
\multicolumn{1}{c|}{\textbf{Target DB}} & \multicolumn{1}{c|}{\textbf{Target FR System}} & \multicolumn{2}{c;{1pt/1pt}}{\textbf{\textit{Single 1}}} & \multicolumn{2}{c;{1pt/1pt}}{\textbf{\textit{Single 2}}} & \multicolumn{2}{c;{1pt/1pt}}{\textbf{\textit{Single 3}}} & \multicolumn{2}{c;{1pt/1pt}}{\textbf{\textit{Single 4}}} & \multicolumn{2}{c|}{\textbf{\textit{Single 5}}} & \multicolumn{2}{c;{1pt/1pt}}{\textbf{\textit{Combi. 1}}} & \multicolumn{2}{c;{1pt/1pt}}{\textbf{\textit{Combi. 2}}} & \multicolumn{2}{c}{\textbf{\textit{Combi. 3}}} \\ 
\hline
\multirow{10}{*}{\begin{tabular}[c]{@{}l@{}}\textbf{LFW -}\\\textbf{ Fold 1}\end{tabular}} & \multirow{2}{*}{\begin{tabular}[c]{@{}l@{}}Inception-ResNet-v2\\ (CASIA-WebFace)\end{tabular}} & \textcolor{blue}{2.3} & \textcolor{blue}{3.3} & 2.3 & 3.3 & {\cellcolor[rgb]{0.753,0.753,0.753}}2.3 & {\cellcolor[rgb]{0.753,0.753,0.753}}3.3 & 2.3 & 3.3 & {\cellcolor[rgb]{0.753,0.753,0.753}}2.3 & {\cellcolor[rgb]{0.753,0.753,0.753}}3.3 & \textcolor{blue}{2.3} & \textcolor{blue}{3.3} & \textcolor{blue}{2.3} & \textcolor{blue}{3.3} & 2.3 & 3.3 \\
 &  & \textbf{\textcolor{blue}{29.9}} & \textbf{\textcolor{blue}{34.7}} & \textbf{15.4} & \textbf{19.7} & {\cellcolor[rgb]{0.753,0.753,0.753}}2.4 & {\cellcolor[rgb]{0.753,0.753,0.753}}2.3 & 0.4 & 0.4 & {\cellcolor[rgb]{0.753,0.753,0.753}}0.8 & {\cellcolor[rgb]{0.753,0.753,0.753}}0.2 & \textbf{\textcolor{blue}{26.6}} & \textbf{\textcolor{blue}{29.7}} & \textbf{\textcolor{blue}{23.4}} & \textbf{\textcolor{blue}{27.3}} & \textbf{5.1} & 4.4 \\ 
\cdashline{2-18}[1pt/1pt]
 & \multirow{2}{*}{\begin{tabular}[c]{@{}l@{}}Inception-ResNet-v2\\ (MS-Celeb)\end{tabular}} & 0.5 & 0.3 & 0.5 & 0.3 & {\cellcolor[rgb]{0.753,0.753,0.753}}\textcolor{blue}{0.5} & {\cellcolor[rgb]{0.753,0.753,0.753}}\textcolor{blue}{0.3} & \textcolor{blue}{0.5} & \textcolor{blue}{0.3} & {\cellcolor[rgb]{0.753,0.753,0.753}}0.5 & {\cellcolor[rgb]{0.753,0.753,0.753}}0.3 & 0.5 & 0.3 & \textcolor{blue}{0.5} & \textcolor{blue}{0.3} & \textcolor{blue}{0.5} & \textcolor{blue}{0.3} \\
 &  & 0.1 & 0.8 & 1.0 & 1.1 & {\cellcolor[rgb]{0.753,0.753,0.753}}\textbf{\textcolor{blue}{7.3}} & {\cellcolor[rgb]{0.753,0.753,0.753}}\textbf{\textcolor{blue}{5.7}} & \textbf{\textcolor{blue}{10.0}} & \textbf{\textcolor{blue}{8.1}} & {\cellcolor[rgb]{0.753,0.753,0.753}}1.1 & {\cellcolor[rgb]{0.753,0.753,0.753}}0.6 & 1.2 & 1.7 & \textcolor{blue}{2.0} & \textcolor{blue}{2.3} & \textcolor{blue}{2.3} & \textcolor{blue}{0.8} \\ 
\cdashline{2-18}[1pt/1pt]
 & \multirow{2}{*}{\begin{tabular}[c]{@{}l@{}}FaceNet (Inception-v1)\\ (MS-Celeb)\end{tabular}} & 0.7 & 0.3 & 0.7 & 0.3 & {\cellcolor[rgb]{0.753,0.753,0.753}}0.7 & {\cellcolor[rgb]{0.753,0.753,0.753}}0.3 & 0.7 & 0.3 & {\cellcolor[rgb]{0.753,0.753,0.753}}0.7 & {\cellcolor[rgb]{0.753,0.753,0.753}}0.3 & 0.7 & 0.3 & 0.7 & 0.3 & 0.7 & 0.3 \\
 &  & 0.0 & 1.1 & 0.3 & 1.1 & {\cellcolor[rgb]{0.753,0.753,0.753}}0.4 & {\cellcolor[rgb]{0.753,0.753,0.753}}0.8 & 0.5 & 1.5 & {\cellcolor[rgb]{0.753,0.753,0.753}}0.4 & {\cellcolor[rgb]{0.753,0.753,0.753}}0.2 & 1.3 & 0.6 & 0.8 & 1.3 & 1.2 & 0.6 \\ 
\cdashline{2-18}[1pt/1pt]
 & \multirow{2}{*}{\begin{tabular}[c]{@{}l@{}}DR-GAN\\ (CASIA-WebFace)\end{tabular}} & 3.3 & 3.7 & \textcolor{blue}{3.3} & \textcolor{blue}{3.7} & {\cellcolor[rgb]{0.753,0.753,0.753}}3.3 & {\cellcolor[rgb]{0.753,0.753,0.753}}3.7 & 3.3 & 3.7 & {\cellcolor[rgb]{0.753,0.753,0.753}}\textcolor{blue}{3.3} & {\cellcolor[rgb]{0.753,0.753,0.753}}\textcolor{blue}{3.7} & \textcolor{blue}{3.3} & \textcolor{blue}{3.7} & 3.3 & 3.7 & \textcolor{blue}{3.3} & \textcolor{blue}{3.7} \\
 &  & \textbf{6.2} & \textbf{8.1} & \textbf{\textcolor{blue}{30.1}} & \textbf{\textcolor{blue}{33.3}} & {\cellcolor[rgb]{0.753,0.753,0.753}}1.1 & {\cellcolor[rgb]{0.753,0.753,0.753}}0.8 & 2.0 & 0.8 & {\cellcolor[rgb]{0.753,0.753,0.753}}\textcolor{blue}{4.3} & {\cellcolor[rgb]{0.753,0.753,0.753}}\textcolor{blue}{3.6} & \textbf{\textcolor{blue}{27.3}} & \textbf{\textcolor{blue}{27.8}} & 3.5 & 2.8 & \textbf{\textcolor{blue}{6.0}} & \textbf{\textcolor{blue}{5.5}} \\ 
\cdashline{2-18}[1pt/1pt]
 & \multirow{2}{*}{\begin{tabular}[c]{@{}l@{}}ArcFace\\ (MS-Celeb)\end{tabular}} & 14.3 & 12.3 & 14.3 & 12.3 & {\cellcolor[rgb]{0.753,0.753,0.753}}14.3 & {\cellcolor[rgb]{0.753,0.753,0.753}}12.3 & 14.3 & 12.3 & {\cellcolor[rgb]{0.753,0.753,0.753}}14.3 & {\cellcolor[rgb]{0.753,0.753,0.753}}12.3 & 14.3 & 12.3 & 14.3 & 12.3 & 14.3 & 12.3 \\
 &  & 11.6 & 13.6 & \textbf{21.3} & \textbf{26.3} & {\cellcolor[rgb]{0.753,0.753,0.753}}2.4 & {\cellcolor[rgb]{0.753,0.753,0.753}}2.5 & 4.9 & 2.5 & {\cellcolor[rgb]{0.753,0.753,0.753}}9.7 & {\cellcolor[rgb]{0.753,0.753,0.753}}7.6 & \textbf{22.1} & \textbf{23.5} & 14.3 & 15.3 & 13.6 & 14.6 \\ 
\hline
\multirow{10}{*}{\textbf{MOBIO}} & \multirow{2}{*}{\begin{tabular}[c]{@{}l@{}}Inception-ResNet-v2\\ (CASIA-WebFace)\end{tabular}} & {\cellcolor[rgb]{0.753,0.753,0.753}}\textcolor{blue}{1.9} & {\cellcolor[rgb]{0.753,0.753,0.753}}\textcolor{blue}{2.1} & {\cellcolor[rgb]{0.753,0.753,0.753}}1.9 & {\cellcolor[rgb]{0.753,0.753,0.753}}2.1 & 1.9 & 2.1 & {\cellcolor[rgb]{0.753,0.753,0.753}}1.9 & {\cellcolor[rgb]{0.753,0.753,0.753}}2.1 & 1.9 & 2.1 & {\cellcolor[rgb]{0.753,0.753,0.753}}\textcolor{blue}{1.9} & {\cellcolor[rgb]{0.753,0.753,0.753}}\textcolor{blue}{2.1} & \textcolor{blue}{1.9} & \textcolor{blue}{2.1} & 1.9 & 2.1 \\
 &  & {\cellcolor[rgb]{0.753,0.753,0.753}}\textcolor{blue}{2.4} & {\cellcolor[rgb]{0.753,0.753,0.753}}\textcolor{blue}{0.0} & {\cellcolor[rgb]{0.753,0.753,0.753}}0.0 & {\cellcolor[rgb]{0.753,0.753,0.753}}0.0 & 0.0 & 0.0 & {\cellcolor[rgb]{0.753,0.753,0.753}}0.0 & {\cellcolor[rgb]{0.753,0.753,0.753}}0.0 & 0.0 & 0.0 & {\cellcolor[rgb]{0.753,0.753,0.753}}\textcolor{blue}{2.4} & {\cellcolor[rgb]{0.753,0.753,0.753}}\textcolor{blue}{0.0} & \textcolor{blue}{0.0} & \textcolor{blue}{0.0} & \textbf{4.8} & \textbf{5.2} \\ 
\cdashline{2-18}[1pt/1pt]
 & \multirow{2}{*}{\begin{tabular}[c]{@{}l@{}}Inception-ResNet-v2\\ (MS-Celeb)\end{tabular}} & {\cellcolor[rgb]{0.753,0.753,0.753}}1.0 & {\cellcolor[rgb]{0.753,0.753,0.753}}0.4 & {\cellcolor[rgb]{0.753,0.753,0.753}}1.0 & {\cellcolor[rgb]{0.753,0.753,0.753}}0.4 & \textcolor{blue}{1.0} & \textcolor{blue}{0.4} & {\cellcolor[rgb]{0.753,0.753,0.753}}\textcolor{blue}{1.0} & {\cellcolor[rgb]{0.753,0.753,0.753}}\textcolor{blue}{0.4} & 1.0 & 0.4 & {\cellcolor[rgb]{0.753,0.753,0.753}}1.0 & {\cellcolor[rgb]{0.753,0.753,0.753}}0.4 & \textcolor{blue}{1.0} & \textcolor{blue}{0.4} & \textcolor{blue}{1.0} & \textcolor{blue}{0.4} \\
 &  & {\cellcolor[rgb]{0.753,0.753,0.753}}0.0 & {\cellcolor[rgb]{0.753,0.753,0.753}}0.0 & {\cellcolor[rgb]{0.753,0.753,0.753}}0.0 & {\cellcolor[rgb]{0.753,0.753,0.753}}0.0 & \textcolor{blue}{0.0} & \textcolor{blue}{0.0} & {\cellcolor[rgb]{0.753,0.753,0.753}}\textcolor{blue}{0.0} & {\cellcolor[rgb]{0.753,0.753,0.753}}\textcolor{blue}{0.0} & 0.0 & 0.0 & {\cellcolor[rgb]{0.753,0.753,0.753}}0.0 & {\cellcolor[rgb]{0.753,0.753,0.753}}0.0 & \textcolor{blue}{0.0} & \textcolor{blue}{0.0} & \textcolor{blue}{0.0} & \textcolor{blue}{0.0} \\ 
\cdashline{2-18}[1pt/1pt]
 & \multirow{2}{*}{\begin{tabular}[c]{@{}l@{}}FaceNet (Inception-v1)\\ (MS-Celeb)\end{tabular}} & {\cellcolor[rgb]{0.753,0.753,0.753}}0.8 & {\cellcolor[rgb]{0.753,0.753,0.753}}0.5 & {\cellcolor[rgb]{0.753,0.753,0.753}}0.8 & {\cellcolor[rgb]{0.753,0.753,0.753}}0.5 & 0.8 & 0.5 & {\cellcolor[rgb]{0.753,0.753,0.753}}0.8 & {\cellcolor[rgb]{0.753,0.753,0.753}}0.5 & 0.8 & 0.5 & {\cellcolor[rgb]{0.753,0.753,0.753}}0.8 & {\cellcolor[rgb]{0.753,0.753,0.753}}0.5 & 0.8 & 0.5 & 0.8 & 0.5 \\
 &  & {\cellcolor[rgb]{0.753,0.753,0.753}}0.0 & {\cellcolor[rgb]{0.753,0.753,0.753}}0.0 & {\cellcolor[rgb]{0.753,0.753,0.753}}0.0 & {\cellcolor[rgb]{0.753,0.753,0.753}}0.0 & 0.0 & 0.0 & {\cellcolor[rgb]{0.753,0.753,0.753}}0.0 & {\cellcolor[rgb]{0.753,0.753,0.753}}0.0 & 0.0 & 1.7 & {\cellcolor[rgb]{0.753,0.753,0.753}}0.0 & {\cellcolor[rgb]{0.753,0.753,0.753}}0.0 & 0.0 & 0.0 & 0.0 & 0.0 \\ 
\cdashline{2-18}[1pt/1pt]
 & \multirow{2}{*}{\begin{tabular}[c]{@{}l@{}}DR-GAN\\ (CASIA-WebFace)\end{tabular}} & {\cellcolor[rgb]{0.753,0.753,0.753}}2.3 & {\cellcolor[rgb]{0.753,0.753,0.753}}1.3 & {\cellcolor[rgb]{0.753,0.753,0.753}}\textcolor{blue}{2.3} & {\cellcolor[rgb]{0.753,0.753,0.753}}\textcolor{blue}{1.3} & 2.3 & 1.3 & {\cellcolor[rgb]{0.753,0.753,0.753}}2.3 & {\cellcolor[rgb]{0.753,0.753,0.753}}1.3 & \textcolor{blue}{2.3} & \textcolor{blue}{1.3} & {\cellcolor[rgb]{0.753,0.753,0.753}}\textcolor{blue}{2.3} & {\cellcolor[rgb]{0.753,0.753,0.753}}\textcolor{blue}{1.3} & 2.3 & 1.3 & \textcolor{blue}{2.3} & \textcolor{blue}{1.3} \\
 &  & {\cellcolor[rgb]{0.753,0.753,0.753}}0.0 & {\cellcolor[rgb]{0.753,0.753,0.753}}0.0 & {\cellcolor[rgb]{0.753,0.753,0.753}}\textcolor{blue}{0.0} & {\cellcolor[rgb]{0.753,0.753,0.753}}\textcolor{blue}{0.0} & 0.0 & 0.0 & {\cellcolor[rgb]{0.753,0.753,0.753}}0.0 & {\cellcolor[rgb]{0.753,0.753,0.753}}0.0 & \textcolor{blue}{2.4} & \textbf{\textcolor{blue}{12.1}} & {\cellcolor[rgb]{0.753,0.753,0.753}}\textcolor{blue}{0.0} & {\cellcolor[rgb]{0.753,0.753,0.753}}\textcolor{blue}{0.0} & 0.0 & 0.0 & \textbf{\textcolor{blue}{4.8}} & \textbf{\textcolor{blue}{6.9}} \\ 
\cdashline{2-18}[1pt/1pt]
 & \multirow{2}{*}{\begin{tabular}[c]{@{}l@{}}ArcFace\\ (MS-Celeb)\end{tabular}} & {\cellcolor[rgb]{0.753,0.753,0.753}}8.2 & {\cellcolor[rgb]{0.753,0.753,0.753}}7.8 & {\cellcolor[rgb]{0.753,0.753,0.753}}8.2 & {\cellcolor[rgb]{0.753,0.753,0.753}}7.8 & 8.2 & 7.8 & {\cellcolor[rgb]{0.753,0.753,0.753}}8.2 & {\cellcolor[rgb]{0.753,0.753,0.753}}7.8 & 8.2 & 7.8 & {\cellcolor[rgb]{0.753,0.753,0.753}}8.2 & {\cellcolor[rgb]{0.753,0.753,0.753}}7.8 & 8.2 & 7.8 & 8.2 & 7.8 \\
 &  & {\cellcolor[rgb]{0.753,0.753,0.753}}0.0 & {\cellcolor[rgb]{0.753,0.753,0.753}}0.0 & {\cellcolor[rgb]{0.753,0.753,0.753}}2.4 & {\cellcolor[rgb]{0.753,0.753,0.753}}1.7 & 0.0 & 0.0 & {\cellcolor[rgb]{0.753,0.753,0.753}}0.0 & {\cellcolor[rgb]{0.753,0.753,0.753}}0.0 & 0.0 & 0.0 & {\cellcolor[rgb]{0.753,0.753,0.753}}4.8 & {\cellcolor[rgb]{0.753,0.753,0.753}}1.7 & 0.0 & 0.0 & 4.8 & 3.4 \\ 
\hline
\multirow{10}{*}{\textbf{IJB-A}} & \multirow{2}{*}{\begin{tabular}[c]{@{}l@{}}Inception-ResNet-v2\\ (CASIA-WebFace)\end{tabular}} & {\cellcolor[rgb]{0.753,0.753,0.753}}\textcolor{blue}{10.1} & {\cellcolor[rgb]{0.753,0.753,0.753}} & {\cellcolor[rgb]{0.753,0.753,0.753}}10.1 & {\cellcolor[rgb]{0.753,0.753,0.753}} & {\cellcolor[rgb]{0.753,0.753,0.753}}10.1 & {\cellcolor[rgb]{0.753,0.753,0.753}} & {\cellcolor[rgb]{0.753,0.753,0.753}}10.1 & {\cellcolor[rgb]{0.753,0.753,0.753}} & {\cellcolor[rgb]{0.753,0.753,0.753}}10.1 & {\cellcolor[rgb]{0.753,0.753,0.753}} & {\cellcolor[rgb]{0.753,0.753,0.753}}\textcolor{blue}{10.1} & {\cellcolor[rgb]{0.753,0.753,0.753}} & {\cellcolor[rgb]{0.753,0.753,0.753}}\textcolor{blue}{10.1} & {\cellcolor[rgb]{0.753,0.753,0.753}} & {\cellcolor[rgb]{0.753,0.753,0.753}}10.1 & {\cellcolor[rgb]{0.753,0.753,0.753}} \\
 &  & {\cellcolor[rgb]{0.753,0.753,0.753}}\textbf{\textcolor{blue}{37.5}} & {\cellcolor[rgb]{0.753,0.753,0.753}} & {\cellcolor[rgb]{0.753,0.753,0.753}}\textbf{15.2} & {\cellcolor[rgb]{0.753,0.753,0.753}} & {\cellcolor[rgb]{0.753,0.753,0.753}}\textbf{13.4} & {\cellcolor[rgb]{0.753,0.753,0.753}} & {\cellcolor[rgb]{0.753,0.753,0.753}}9.8 & {\cellcolor[rgb]{0.753,0.753,0.753}} & {\cellcolor[rgb]{0.753,0.753,0.753}}2.7 & {\cellcolor[rgb]{0.753,0.753,0.753}} & {\cellcolor[rgb]{0.753,0.753,0.753}}\textbf{\textcolor{blue}{20.5}} & {\cellcolor[rgb]{0.753,0.753,0.753}} & {\cellcolor[rgb]{0.753,0.753,0.753}}\textbf{\textcolor{blue}{25.0}} & {\cellcolor[rgb]{0.753,0.753,0.753}} & {\cellcolor[rgb]{0.753,0.753,0.753}}2.7 & {\cellcolor[rgb]{0.753,0.753,0.753}} \\ 
\cdashline{2-18}[1pt/1pt]
 & \multirow{2}{*}{\begin{tabular}[c]{@{}l@{}}Inception-ResNet-v2\\ (MS-Celeb)\end{tabular}} & {\cellcolor[rgb]{0.753,0.753,0.753}}3.1 & {\cellcolor[rgb]{0.753,0.753,0.753}} & {\cellcolor[rgb]{0.753,0.753,0.753}}3.1 & {\cellcolor[rgb]{0.753,0.753,0.753}} & {\cellcolor[rgb]{0.753,0.753,0.753}}\textcolor{blue}{3.1} & {\cellcolor[rgb]{0.753,0.753,0.753}} & {\cellcolor[rgb]{0.753,0.753,0.753}}\textcolor{blue}{3.1} & {\cellcolor[rgb]{0.753,0.753,0.753}} & {\cellcolor[rgb]{0.753,0.753,0.753}}3.1 & {\cellcolor[rgb]{0.753,0.753,0.753}} & {\cellcolor[rgb]{0.753,0.753,0.753}}3.1 & {\cellcolor[rgb]{0.753,0.753,0.753}} & {\cellcolor[rgb]{0.753,0.753,0.753}}\textcolor{blue}{3.1} & {\cellcolor[rgb]{0.753,0.753,0.753}} & {\cellcolor[rgb]{0.753,0.753,0.753}}\textcolor{blue}{3.1} & {\cellcolor[rgb]{0.753,0.753,0.753}} \\
 &  & {\cellcolor[rgb]{0.753,0.753,0.753}}0.0 & {\cellcolor[rgb]{0.753,0.753,0.753}} & {\cellcolor[rgb]{0.753,0.753,0.753}}1.8 & {\cellcolor[rgb]{0.753,0.753,0.753}} & {\cellcolor[rgb]{0.753,0.753,0.753}}\textbf{\textcolor{blue}{20.5}} & {\cellcolor[rgb]{0.753,0.753,0.753}} & {\cellcolor[rgb]{0.753,0.753,0.753}}\textbf{\textcolor{blue}{22.3}} & {\cellcolor[rgb]{0.753,0.753,0.753}} & {\cellcolor[rgb]{0.753,0.753,0.753}}0.0 & {\cellcolor[rgb]{0.753,0.753,0.753}} & {\cellcolor[rgb]{0.753,0.753,0.753}}1.8 & {\cellcolor[rgb]{0.753,0.753,0.753}} & {\cellcolor[rgb]{0.753,0.753,0.753}}\textcolor{blue}{0.9} & {\cellcolor[rgb]{0.753,0.753,0.753}} & {\cellcolor[rgb]{0.753,0.753,0.753}}\textcolor{blue}{0.0} & {\cellcolor[rgb]{0.753,0.753,0.753}} \\ 
\cdashline{2-18}[1pt/1pt]
 & \multirow{2}{*}{\begin{tabular}[c]{@{}l@{}}FaceNet (Inception-v1)\\ (MS-Celeb)\end{tabular}} & {\cellcolor[rgb]{0.753,0.753,0.753}}5.7 & {\cellcolor[rgb]{0.753,0.753,0.753}} & {\cellcolor[rgb]{0.753,0.753,0.753}}5.7 & {\cellcolor[rgb]{0.753,0.753,0.753}} & {\cellcolor[rgb]{0.753,0.753,0.753}}5.7 & {\cellcolor[rgb]{0.753,0.753,0.753}} & {\cellcolor[rgb]{0.753,0.753,0.753}}5.7 & {\cellcolor[rgb]{0.753,0.753,0.753}} & {\cellcolor[rgb]{0.753,0.753,0.753}}5.7 & {\cellcolor[rgb]{0.753,0.753,0.753}} & {\cellcolor[rgb]{0.753,0.753,0.753}}5.7 & {\cellcolor[rgb]{0.753,0.753,0.753}} & {\cellcolor[rgb]{0.753,0.753,0.753}}5.7 & {\cellcolor[rgb]{0.753,0.753,0.753}} & {\cellcolor[rgb]{0.753,0.753,0.753}}5.7 & {\cellcolor[rgb]{0.753,0.753,0.753}} \\
 &  & {\cellcolor[rgb]{0.753,0.753,0.753}}4.5 & {\cellcolor[rgb]{0.753,0.753,0.753}} & {\cellcolor[rgb]{0.753,0.753,0.753}}2.7 & {\cellcolor[rgb]{0.753,0.753,0.753}} & {\cellcolor[rgb]{0.753,0.753,0.753}}\textbf{9.8} & {\cellcolor[rgb]{0.753,0.753,0.753}} & {\cellcolor[rgb]{0.753,0.753,0.753}}\textbf{8.0} & {\cellcolor[rgb]{0.753,0.753,0.753}} & {\cellcolor[rgb]{0.753,0.753,0.753}}1.8 & {\cellcolor[rgb]{0.753,0.753,0.753}} & {\cellcolor[rgb]{0.753,0.753,0.753}}6.2 & {\cellcolor[rgb]{0.753,0.753,0.753}} & {\cellcolor[rgb]{0.753,0.753,0.753}}4.5 & {\cellcolor[rgb]{0.753,0.753,0.753}} & {\cellcolor[rgb]{0.753,0.753,0.753}}4.5 & {\cellcolor[rgb]{0.753,0.753,0.753}} \\ 
\cdashline{2-18}[1pt/1pt]
 & \multirow{2}{*}{\begin{tabular}[c]{@{}l@{}}DR-GAN\\ (CASIA-WebFace)\end{tabular}} & {\cellcolor[rgb]{0.753,0.753,0.753}}10.8 & {\cellcolor[rgb]{0.753,0.753,0.753}} & {\cellcolor[rgb]{0.753,0.753,0.753}}\textcolor{blue}{10.8} & {\cellcolor[rgb]{0.753,0.753,0.753}} & {\cellcolor[rgb]{0.753,0.753,0.753}}10.8 & {\cellcolor[rgb]{0.753,0.753,0.753}} & {\cellcolor[rgb]{0.753,0.753,0.753}}10.8 & {\cellcolor[rgb]{0.753,0.753,0.753}} & {\cellcolor[rgb]{0.753,0.753,0.753}}\textcolor{blue}{10.8} & {\cellcolor[rgb]{0.753,0.753,0.753}} & {\cellcolor[rgb]{0.753,0.753,0.753}}\textcolor{blue}{10.8} & {\cellcolor[rgb]{0.753,0.753,0.753}} & {\cellcolor[rgb]{0.753,0.753,0.753}}10.8 & {\cellcolor[rgb]{0.753,0.753,0.753}} & {\cellcolor[rgb]{0.753,0.753,0.753}}\textcolor{blue}{10.8} & {\cellcolor[rgb]{0.753,0.753,0.753}} \\
 &  & {\cellcolor[rgb]{0.753,0.753,0.753}}4.5 & {\cellcolor[rgb]{0.753,0.753,0.753}} & {\cellcolor[rgb]{0.753,0.753,0.753}}\textbf{\textcolor{blue}{16.1}} & {\cellcolor[rgb]{0.753,0.753,0.753}} & {\cellcolor[rgb]{0.753,0.753,0.753}}7.1 & {\cellcolor[rgb]{0.753,0.753,0.753}} & {\cellcolor[rgb]{0.753,0.753,0.753}}\textbf{14.3} & {\cellcolor[rgb]{0.753,0.753,0.753}} & {\cellcolor[rgb]{0.753,0.753,0.753}}\textcolor{blue}{5.4} & {\cellcolor[rgb]{0.753,0.753,0.753}} & {\cellcolor[rgb]{0.753,0.753,0.753}}\textbf{\textcolor{blue}{17.9}} & {\cellcolor[rgb]{0.753,0.753,0.753}} & {\cellcolor[rgb]{0.753,0.753,0.753}}5.4 & {\cellcolor[rgb]{0.753,0.753,0.753}} & {\cellcolor[rgb]{0.753,0.753,0.753}}\textcolor{blue}{1.8} & {\cellcolor[rgb]{0.753,0.753,0.753}} \\ 
\cdashline{2-18}[1pt/1pt]
 & \multirow{2}{*}{\begin{tabular}[c]{@{}l@{}}ArcFace\\ (MS-Celeb)\end{tabular}} & {\cellcolor[rgb]{0.753,0.753,0.753}}10.7 & {\cellcolor[rgb]{0.753,0.753,0.753}} & {\cellcolor[rgb]{0.753,0.753,0.753}}10.7 & {\cellcolor[rgb]{0.753,0.753,0.753}} & {\cellcolor[rgb]{0.753,0.753,0.753}}10.7 & {\cellcolor[rgb]{0.753,0.753,0.753}} & {\cellcolor[rgb]{0.753,0.753,0.753}}10.7 & {\cellcolor[rgb]{0.753,0.753,0.753}} & {\cellcolor[rgb]{0.753,0.753,0.753}}10.7 & {\cellcolor[rgb]{0.753,0.753,0.753}} & {\cellcolor[rgb]{0.753,0.753,0.753}}10.7 & {\cellcolor[rgb]{0.753,0.753,0.753}} & {\cellcolor[rgb]{0.753,0.753,0.753}}10.7 & {\cellcolor[rgb]{0.753,0.753,0.753}} & {\cellcolor[rgb]{0.753,0.753,0.753}}10.7 & {\cellcolor[rgb]{0.753,0.753,0.753}} \\
 &  & {\cellcolor[rgb]{0.753,0.753,0.753}}0.0 & {\cellcolor[rgb]{0.753,0.753,0.753}} & {\cellcolor[rgb]{0.753,0.753,0.753}}3.6 & {\cellcolor[rgb]{0.753,0.753,0.753}} & {\cellcolor[rgb]{0.753,0.753,0.753}}0.9 & {\cellcolor[rgb]{0.753,0.753,0.753}} & {\cellcolor[rgb]{0.753,0.753,0.753}}0.9 & {\cellcolor[rgb]{0.753,0.753,0.753}} & {\cellcolor[rgb]{0.753,0.753,0.753}}2.7 & {\cellcolor[rgb]{0.753,0.753,0.753}} & {\cellcolor[rgb]{0.753,0.753,0.753}}0.9 & {\cellcolor[rgb]{0.753,0.753,0.753}} & {\cellcolor[rgb]{0.753,0.753,0.753}}0.9 & {\cellcolor[rgb]{0.753,0.753,0.753}} & {\cellcolor[rgb]{0.753,0.753,0.753}}0.9 & {\cellcolor[rgb]{0.753,0.753,0.753}} \\
\hline
\end{tabular}}
\end{table*}

Table~\ref{tab:fmrs} shows the FMRs of normal tests and the corresponding master face tests using master faces generated using five \textit{single} settings and three \textit{combination} settings. Each cell has 4 numbers and they are the FMRs of the normal tests (upper part) and the master face tests (lower part), from the development sets (left) and the evaluation sets of the target databases (right), respectively. Gray cells indicate the surrogate database(s) used by attackers when running the LVE algorithm and the target database(s) are different, while gray cells imply that they are the same. Numbers in bold indicate success master face attacks.
There are several observations regarding the FMRs of the attacks using the master faces generated using the \textit{single} and \textit{combination} settings shown in Table~\ref{tab:fmrs} in connection with the FMR curves shown in Fig.~\ref{fig:fmr} discussed above:

\begin{itemize}
\item All FR systems are vulnerable to master face attacks. Some systems are easier to fool than others.
\item With the \textit{combination 1} setting, the master face had the attack abilities of the master faces generated using the corresponding single settings (\textit{single 1} and \textit{single 2}). In this case, there was no conflict.
\item With the \textit{combination 2} and \textit{combination 3} settings in which conflict occurred, their master faces were lacking some attack abilities of the master faces generated using the corresponding single settings. This is clearly seen for the \textit{combination 2} setting, for which six attacks which were success in the single settings were failed.
\item With the \textit{combination 3} setting, for which the two component databases and FR systems differed, besides five lost attacks, there were six new successful ones. Moreover, the FMRs of the successful attacks of the combination setting were not as high as the those of the single setting. Conflict still occurred in this case, however, it is less severe than in the \textit{combination 2} and it caused the changes in the successful cases.
\end{itemize}

\begin{table}[t!]
    \centering
    \caption{Summary of successful attack ratios using \textbf{five} \textit{single} settings and \textbf{three} combination settings. The numerators are the numbers of successful attacks and the denominators are the total numbers of attack cases.}
	\label{tab:summary}
	\adjustbox{max width=\columnwidth}{
		\begin{tabular}{l|c|c|cc}
		    \hline
			\multicolumn{1}{c|}{\multirow{2}{*}{\textbf{FR Setting}}} & \multicolumn{2}{c|}{\textbf{\textit{Single} Settings}} & \multicolumn{2}{c}{\textbf{\textit{Combination} Settings}} \\ \cline{2-5} 
			\multicolumn{1}{l|}{} & \multicolumn{1}{c|}{\textbf{\begin{tabular}[c]{@{}c@{}}Known\\ DB\end{tabular}}} & \multicolumn{1}{c|}{\textbf{\begin{tabular}[c]{@{}c@{}}Unknown\\ DB\end{tabular}}} & \multicolumn{1}{c|}{\textbf{\begin{tabular}[c]{@{}c@{}}Known\\ DB\end{tabular}}} & \multicolumn{1}{c}{\textbf{\begin{tabular}[c]{@{}c@{}}Unknown\\ DB\end{tabular}}} \\ \hline
			Same Arch. - Same DB & \textbf{9/12} & \textbf{7/24} & \multicolumn{1}{c|}{\textbf{10/20}} & \textbf{3/16} \\
			Same Arch. - Different DB & 0/8 & \textbf{1/16} & \multicolumn{1}{c|}{\textbf{3/6}} & 0/6 \\
			Different Arch. - Same DB & \textbf{6/16} & \textbf{3/32} & \multicolumn{1}{c|}{0/20} & 0/10 \\
			Different Arch. - Different DB & \textbf{2/24} & \textbf{1/48} & \multicolumn{1}{c|}{\textbf{2/4}} & 0/8 \\ \hline
	\end{tabular}}
\end{table}

Table~\ref{tab:summary} summarizes the number of successful attacks using both \textit{single} and \textit{combination} settings. An attack is successful if the master face's FMR is higher than the normal test set's FMR. Recall that there were \textbf{five} \textit{single} settings and \textbf{three} \textit{combination} settings. Moreover, the \textit{combination} settings used more than one database and/or one FR system, and there were only three databases and five FR systems used for evaluation. As a result, the number of successful black/gray-box attacks (attacks on different architecture, different database) with these settings was less than that of attacks with the \textit{single} settings.

The results from this experiment provide valuable clues for effectively designing the LVE algorithm. Using only one database is a safe way to avoid conflicts. Besides conflicts, using both different databases and different FR systems may result in some unpredictable good results in some cases. This setting may be useful when performing black-box or gray-box attacks when the setting using only one database failed.

\section{Presentation Attacks}
\label{sec:presentation_attacks}
Finally, we evaluated the risk and threats of presentation attacks using master faces on the FR systems. For master face candidates, we chose one generated using the \textit{single 2} setting and another generated using the \textit{combination 1} setting. For digital attack candidates, we chose two attack scenarios on the IJB-A database~\cite{klare2015pushing} in which the two master faces were falsely accepted by the Inception-ResNet-v2 based FR system~\cite{de2018heterogeneous} (CASIA-WebFace version) and the DR-GAN FR system~\cite{tran2017disentangled}. We compared the FMRs of these two digital attacks with those of the corresponding presentation attacks.

\subsection{Experiment design}

\begin{figure}[th!]
	\centering
	\includegraphics[width=70mm]{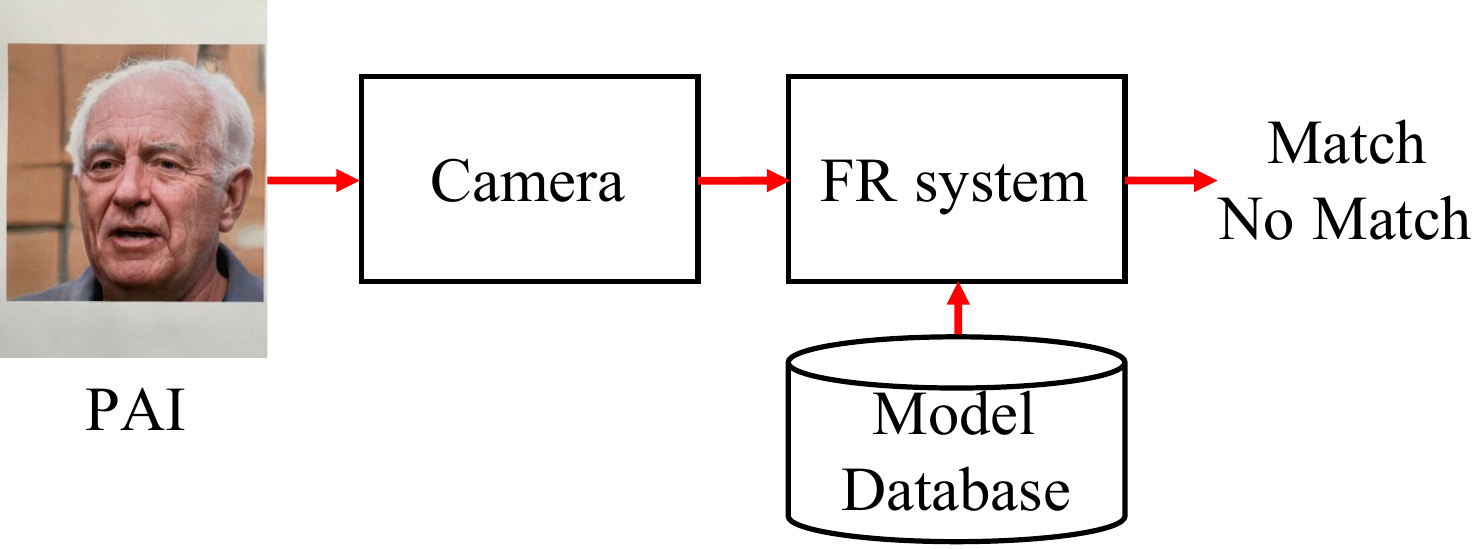}
	\caption{Overview of presentation attack on FR system.}
	\label{fig:presentation_attack}
\end{figure}

To simulate simple presentation attacks like the one shown in Fig.~\ref{fig:presentation_attack}, we needed to prepare PAIs and cameras. For the PAIs of each of the two selected master faces, we used three kinds of materials:
\begin{itemize}
\item Color photos printed on plain A4 paper.
\item Color photos printed on 127 mm $\times$ 178 mm photo paper.
\item Color photos displayed on the screen of an Apple 13-inch MacBook Pro 2017.
\end{itemize}

For the cameras, we used two types:
\begin{itemize}
\item the rear camera in an iPhone XR.
\item a Canon EOS 60D DSLR camera with a Canon EF 40mm F2.8 STM lens.
\end{itemize}

For simplicity, we used these cameras to take photos of the PAIs under normal room conditions. We adjusted the position of the cameras such that they were relatively perpendicular to the surface of the PAIs so they could capture the displayed PAIs as much as possible without loosing any contents. This condition is close to that of real-world presentation attacks. Three example PAIs are shown in Fig.~\ref{fig:masterfaces}.

\subsection{Results}

\begin{table}[th!]
    \centering
    \caption{FMRs of master face PAI attacks on dev set of IJB-A database~\cite{klare2015pushing} using two settings: \textit{Single 2} and \textit{combination 1}. First line in each row shows result for Inception-ResNet-v2 based FR system~\cite{de2018heterogeneous} trained on CASIA-WebFace database, and second line shows result for DR-GAN FR system~\cite{tran2017disentangled}. Numbers in \textbf{bold font} indicate successful attacks although there was degradation in the FMR in some cases. Numbers in \textcolor{red}{\textbf{red}} indicate the increment of the FMRs in the presentation attack compared with those in digital attacks. }
    \label{tab:presentation_attacks}
    \adjustbox{max width=\columnwidth}{
    \begin{tabular}{l|ccc|cc}
        \hline
        \multicolumn{1}{c|}{\textbf{Camera}} & \textbf{Plain Paper} & \textbf{Photo Paper} & \begin{tabular}[c]{@{}c@{}}\textbf{MacBook}\\ \textbf{Screen}\end{tabular} & \begin{tabular}[c]{@{}c@{}}\textbf{Digital}\\ \textbf{Master} \\ \textbf{Face}\end{tabular} & \begin{tabular}[c]{@{}c@{}}\textbf{Normal}\\ \textbf{Dev Set}\end{tabular} \\ \hline
        & \multicolumn{4}{l}{Setting: \textit{Single 2}} \\
        \hdashline
        \multirow{2}{*}{iPhone XR} & \textbf{17.9 \textcolor{red}{($+$2.7)}} & \textbf{15.2 ($\;\;\;$0.0)} & \textbf{13.4 ($-$1.8)} & 15.2 & 10.1 \\
        & \textbf{18.8 \textcolor{red}{($+$2.7)}} & 11.6 ($-$4.5) & \textbf{19.6 \textcolor{red}{($+$3.5)}} & 16.1 & 10.8 \\
        \hdashline
        \multirow{2}{*}{Canon 60D} & \textbf{17.9 \textcolor{red}{($+$2.7)}} & \textbf{18.8 \textcolor{red}{($+$3.6)}} & \textbf{19.6 \textcolor{red}{($+$4.4)}} & 15.2 & 10.1 \\
        & \textbf{20.5 \textcolor{red}{($+$4.4)}} & 11.6 ($-$4.5) & \textbf{15.2 ($-$0.9)} & 16.1 & 10.8 \\ \hline
        & \multicolumn{5}{l}{Setting: \textit{Combination 1}} \\
        \hdashline
        \multirow{2}{*}{iPhone XR} & \textbf{18.8 ($-$1.7)} & \textbf{19.6 ($-$0.9)} & \textbf{15.2 ($-$5.3)} & 20.5 & 10.1 \\
        & \textbf{13.4 ($-$4.5)} & 11.6 ($-$6.3) & $\;\;$8.9 ($-$9.0) & 17.9 & 10.8 \\
        \hdashline
        \multirow{2}{*}{Canon 60D} & \textbf{17.9 ($-$2.6)} & \textbf{19.6 ($-$0.9)} & \textbf{20.5 ($\;\;\;$0.0)} & 20.5 & 10.1 \\
        & \textbf{13.4 ($-$4.5)} & $\;\;$8.0 ($-$9.9) & \textbf{18.8 \textcolor{red}{($+$0.9)}} & 17.9 & 10.8 \\ \hline
    \end{tabular}}
\end{table}

The FMRs of the attacks using PAI master faces are shown in Table~\ref{tab:presentation_attacks} along with those of attacks using digital master faces and those of the normal dev set of the IJB-A database. The attacks were successful in 19 of the 24 cases, demonstrating that PAI master faces can be effective in real-world attacks. In eight cases, the FMRs were higher than those of attacks using digital master faces. This is attributed to the distribution of PAI master faces being closer to the distribution of faces in the facial databases (which contain faces also captured with a camera) thanks to the camera processing. The lower rate in the other cases is attributed to artifacts from the PAI materials playing a bigger role than the effect of the camera processing. All of the PAI attacks using plain paper were successful while seven of the eight PAI attacks using a computer screen were successful. The attacks using photo paper, which easily reflects light had the worst performance. Those using photos taken with the iPhone camera were more successful than those using ones taken with the Canon camera. This is attributed to the Canon camera being able to capture more detailed PAI artifacts.

\section{Defense Against Master Face Attacks}
\label{sec:defense}
What is the main problem of existing FR systems that causes the existence of master faces? We hypothesized that it comes from the distributions of the embedding spaces where the extracted features are not well distributed. This results in the formation of clusters, not only multi-identity clusters but also age and gender ones.
There are two possible origins of this problem: (1) the training data and (2) the objective function design. Regarding (1), as shown in Figs.~\ref{fig:age} and~\ref{fig:gender}, the training data was unbalanced in terms of age and gender. This could affect the distribution of the embeddings for which the FR systems discriminate well on the majority group than the minority one. For example, the 30-60 year-old face embeddings were scattered more uniformly than the others, as shown in Fig~\ref{fig:density}. Simply increasing the database size has a certain effect on the robustness of the FR systems (the MS-Celeb version of the Inception-ResNet-v2 based FR system had fewer successful master face attacks than the CASIA-WebFace version); however, they are still vulnerable. It is thus important to balance the training data.
Regarding (2), the objective functions are mainly designed so that same-identity embeddings stay close together while different-identity ones stay far apart. The introduction of the angular margin loss~\cite{deng2019arcface} improves this ability while the uniform loss~\cite{duan2019uniformface} forces the embeddings to be uniformly distributed. Although these improvements reduce the risk of master face attacks, they mainly focus on identity. Since gender, age, and race are also important~\cite{grother2019face}, the attack is successful in some cases. This suggests that the design of the objective functions used for training the FR systems needs further improvement.

Beside harnessing FR systems, using master face detectors could mitigate master face attacks. Since master faces are generated using GAN, some GAN image detectors~\cite{marra2019incremental, yu2019attributing, hulzebosch2020detecting} or deepfake detectors~\cite{tolosana2020deepfakes} could be used to detect them.
Although looking realistic from the human perspective, computer-generated images have different properties than natural ones captured by cameras. Some GAN artifacts may exist in the generated images; therefore, most GAN image detectors focus on detecting their presence. We could also integrate a presentation attack detector~\cite{tolosana2020deepfakes} with an FR system to prevent master face attacks as well as other traditional presentation attacks using images or videos of the victims. However, generalization of these detectors is still a huge challenge. The StyleGAN used in the LVE algorithm could be replaced with a more advanced facial generator to fool fake image detectors. Although some degree of generalizability has been achieved, performance is still not good enough for real-world applications. Therefore, further research on generalizability is needed.

\section{Conclusion}
\label{sec:conclusion}
We have again demonstrated, especially in our presentation attack experiment, that master face attacks pose a significant security threat if the FR systems are not properly protected. Our intensive evaluation of the performance of the LVE algorithm using several settings including both \textit{single} and \textit{combination} settings has brought to light several properties of master faces as well as of the LVE algorithm. Some of the \textit{combination} settings caused intra-component conflicts while others produced interesting positive results. Being aware of the existence of master faces and their properties is critical to improving the robustness of face recognition systems. Combining the use of a face recognition system with a well-designed objective function trained on a large balanced database with a fake image detector could mitigate master face attacks. Since digital attack detectors (GAN image detectors and deepfake detectors) and presentation attack detectors still have difficulty with generalization and master face attacks continue to improve, these attacks cannot be taken lightly. Future work will focus on designing a better method to generate master faces and detecting master face attacks.

\section*{Acknowledgments}
This work was partially supported by JSPS KAKENHI Grants JP16H06302, JP18H04120, JP21H04907, JP20K23355, and JP21K18023, and by JST CREST Grants JPMJCR18A6 and JPMJCR20D3, including the AIP challenge program, Japan.

We would like to thank Dr. Tiago de Freitas Pereira and Dr. Amir Mohammadi of the Biometrics Security and Privacy (BSP) group at the Idiap Research Institute for providing the pre-trained face recognition systems and for their support with the Bob toolkit.

\ifCLASSOPTIONcaptionsoff
  \newpage
\fi

\bibliographystyle{IEEEtran}
\bibliography{refs}

\begin{IEEEbiographynophoto}{Huy H. Nguyen} received a B.S. degree in Information Technology from VNUHCM - University of Science, Ho Chi Minh City, Vietnam in 2013. He is currently pursuing a Ph.D. degree in computer science at the Graduate University for Advanced Studies (SOKENDAI), Kanagawa, Japan. His current research interests include security and privacy in biometrics and machine learning.
\end{IEEEbiographynophoto}

\begin{IEEEbiographynophoto}{S\'ebastien Marcel}
is the head of the Biometrics Security and Privacy (BSP) group at the Idiap Research Institute (CH) and conducts research on face recognition, speaker recognition, vein recognition and presentation attack detection. He is a lecturer at the {\'E}cole Polytechnique F{\'e}d{\'e}rale de Lausanne and the University of Lausanne. He is Associate Editor of IEEE Transactions on Biometrics, Behavior, and Identity Science (TBIOM). He was the coordinator of European research projects including MOBIO, TABULA RASA or BEAT and involved in international projects (DARPA, IARPA). He is also the Director of the Swiss Center for Biometrics Research and Testing conducting FIDO certifications and research.

\end{IEEEbiographynophoto}

\begin{IEEEbiographynophoto}{Junichi Yamagishi} (SM’13) received the Ph.D.\ degree from the Tokyo Institute of Technology (Tokyo Tech), Tokyo, Japan, in 2006. From 2007-2013 he was a research fellow in the Centre for Speech Technology Research (CSTR) at the University of Edinburgh, UK. He was appointed Associate Professor at National Institute of Informatics, Japan in 2013. He is currently a Professor at NII, Japan. His research topics include speech processing, machine learning, signal processing, biometrics, digital media cloning and media forensics. 

He served previously as co-organizer for the bi-annual ASVspoof challenge and the bi-annual Voice conversion challenge. He also served as a member of the IEEE Speech and Language Technical Committee (2013-2019), an Associate Editor of the IEEE/ACM Transactions on Audio Speech and Language Processing (2014-2017), and a chairperson of ISCA SynSIG (2017- 2021). He is currently a PI of JST-CREST and ANR supported VoicePersona project and a Senior Area Editor of the IEEE/ACM TASLP.
\end{IEEEbiographynophoto}

\begin{IEEEbiographynophoto}{Isao Echizen} received B.S., M.S., and D.E. degrees from the Tokyo Institute of Technology, Japan, in 1995, 1997, and 2003, respectively. He joined Hitachi, Ltd. in 1997 and until 2007 was a research engineer in the company's systems development laboratory. He is currently a director and a professor of the Information and Society Research Division, the National Institute of Informatics (NII), and a professor in the Department of Information and Communication Engineering, Graduate School of Information Science and Technology, The University of Tokyo, Japan. He was a visiting professor at Tsuda University, Japan, at the University of Freiburg, Germany, and at the University of Halle-Wittenberg, Germany. He is currently engaged in research on multimedia security and multimedia forensics. He currently serves as a research director of the CREST FakeMedia project, Japan Science and Technology Agency (JST). He was a member of the Information Forensics and Security Technical Committee and the IEEE Signal Processing Society. He is the Japanese representative on IFIP TC11 (Security and Privacy Protection in Information Processing Systems), a member-at-large of the APSIPA Board of Governors, and an editorial board member of the IEEE Transactions on Dependable and Secure Computing and the EURASIP Journal on Image and Video Processing.
\end{IEEEbiographynophoto}

\end{document}